\documentclass[12pt,psfig]{article}
\usepackage{graphicx}
\usepackage{float}
\usepackage{subcaption}
\usepackage{amsmath,amssymb,mathtools,bbm}
\usepackage{bbold}
\usepackage{color}
\usepackage{tikz}
\usetikzlibrary{shapes}
\usepackage{booktabs}
\usepackage{multirow}
\usepackage{longtable}
\usepackage{pgfplots}
\usepackage{subeqnarray}
\usepackage{dsfont}
\usepackage{changepage}
\usepackage{hyperref}

\tikzset{cross/.style={cross out, draw=black, minimum size=2*(#1-\pgflinewidth), inner sep=0pt, outer sep=0pt},
cross/.default={1pt}}

\usepackage[symbol]{footmisc} %footnote symbol

\usepackage[bibstyle= nature, citestyle=numeric-comp, sorting=none]{biblatex}

\usepackage{cleveref}

\begin{document} 

\begin{center}

\begin{Large} 
\noindent Application of deep learning to large scale riverine flow velocity estimation \\
\end{Large}

\vspace{0.7cm}
\begin{small}
\noindent Mojtaba Forghani$^1$\footnote[1]{email: mojtaba@stanford.edu}, Yizhou Qian$^2$, Jonghyun Lee$^3$, Matthew Farthing$^4$, Tyler Hesser$^4$, Peter K\@. Kitanidis$^{2,5}$, and Eric F\@. Darve$^{1, 2}$ \\
\end{small}

\vspace{0.3cm}
\begin{footnotesize}
$^1$ Department of Mechanical Engineering, Stanford University, CA \\
$^2$ Institute for Computational and Mathematical Engineering, Stanford University, CA\\
$^3$ Department of Civil and Environmental Engineering and Water Resources Research Center, University of Hawaii at Manoa, Honolulu, HI \\
$^4$ U.S\@. Army Engineer Research and Development Center, Vicksburg, MS\\
$^5$ Department of Civil and Environmental Engineering, Stanford University, CA
\end{footnotesize}

\end{center}

\begin{abstract}
Fast and reliable prediction of riverine flow velocities plays an important role in many applications, including flood risk management. The shallow water equations (SWEs) are commonly used for prediction of the riverine flow velocities. However, accurate and fast prediction with standard SWE solvers remains challenging in many cases. Traditional approaches are computationally expensive and require high-resolution riverbed profile measurement (i.e., bathymetry) for accurate predictions. As a result, they are a poor fit in situations where they need to be evaluated repetitively due, for example, to varying boundary condition (BC) scenarios, or when the bathymetry is not known with certainty. In this work, we propose a two-stage process that tackles these issues. First, using the principal component geostatistical approach (PCGA) we estimate the probability density function of the bathymetry from flow velocity measurements, and then we use multiple machine learning algorithms in order to obtain a fast solver of the SWEs, given augmented realizations from the posterior bathymetry distribution and the prescribed range of potential BCs. The first step of the proposed approach allows us to predict flow velocities without any direct measurement of the bathymetry. Furthermore, the augmentation of the distribution in the second stage allows incorporation of the additional bathymetry information into the flow velocity prediction for improved accuracy and generalization, even if the bathymetry changes over time. Here, we use three different forward solvers, referred to as PCA-DNN (principal component analysis-deep neural network), SE (supervised encoder), and SVE (supervised variational encoder), and validate them on a reach of the Savannah river near Augusta, GA. Our results show that the fast solvers are capable of predicting flow velocities with variable bathymetry and BCs with good accuracy, at a computational cost that is significantly lower than the cost of solving the full boundary value problem with traditional methods.
\end{abstract}

{\it keywords}: deep learning, riverine bathymetry, riverine flow velocity, shallow water equations, reduced order models

\section{Introduction}
\label{intro}

Estimation of riverine flow velocities plays an essential role in many practical applications such as the study of river morphodynamics, safe and efficient maritime transportation, and flood risk management~\cite{overdeep, morph, Casas, Westaway, Lane}. In order to accurately estimate flow velocities under variable boundary conditions (BCs), such as the discharge and the free-surface elevation, as well as the riverbed profile, also known as the bathymetry, we require an accurate predictor of the flow velocities given the bathymetry and the BCs. The shallow water equations (SWEs) are typically used to solve this problem~\cite{Landon, Wilson, Antuono, Xia, Cea, Singh, Horvath}. However, other than a few simple cases such as idealized one-dimensional (1D) flow~\cite{Novikov} or idealized hyperbolic riverbed profiles~\cite{Matskevich}, these equations must be solved numerically~\cite{Benkhaldoun, Delis, Lee_SPH}. Thus they can be computationally expensive and even prohibitive in cases where the simulations need to be run in resource-limited conditions or even on standard desktop configurations, unless we have access to fast graphics processing units (GPUs) and special purpose solvers~\cite{shallow_GPU}. The high computational cost of common numerical solvers of the SWEs is an important shortcoming of these methods, since BCs in rivers can vary widely and thus having a ``fast online predictor'' of the flow velocities is very important, in particular, in situations when a range of conditions need to be evaluated quickly to address questions related to navigability or to asses the risk of flooding.

Other than the computational challenges, SWE solvers typically require a fairly high resolution of the bathymetry as simulation input in order to have accurate prediction of flow velocities. However, direct high-resolution bathymetric surveys by wading or watercraft-mounted multibeam sonar equipment~\cite{Casas, Marcus_mapping} are time consuming and costly for long river reaches. Therefore, combining remote sensing techniques that monitor bathymetry changes in the rivers without direct measurement~\cite{Emery, Inference} with a fast solver of the SWEs offers a valuable tool for flow velocity estimation under variable BCs and bathymetries. In this work, we propose a two-stage process in which, first, the river bathymetry at a site of interest is estimated with uncertainty quantification using indirect remote sensing or drift observations, and then the distribution of estimated bathymetry is augmented and combined with different BCs to obtain a fast solver of the SWEs, that is, a predictor of depths and flow velocities for a range of possible BCs that may arise in future conditions or alternative planning scenarios. Note that in the following we focus on the flow velocities for simplicity. Once the solver is trained (at the offline stage), the prediction (at the online stage) can be performed multiple orders of magnitude faster than common numerical solvers. 

When updated bathymetry measurements are not available, our solver is capable of predicting a flow velocity distribution consistent with the estimated bathymetry posterior distribution, that is, mean and the standard deviation of flow velocity for different BCs. Furthermore, if bathymetry measurements at a limited number of cross sections (or all cross sections) over the riverine domain become available, this information can be incorporated into the velocity prediction without any further training---and thus can be done at no additional computational cost. Below, we will demonstrate that the inclusion of the incomplete bathymetry measurements into the prediction has the potential to increase the accuracy of the flow velocity predictions in the regions with the bathymetry measurement, without reducing the quality of the prediction in regions where no bathymetry measurement is available. 

Stage one of the proposed two-stage process requires, first, a measurable flow property from which we can obtain an estimate of the bathymetry, and second, a numerical approach that can be used to reconstruct bathymetry from the information contained in the measured flow property. Several remote sensing techniques have been used in the literature to obtain informative flow properties from which the bathymetry can be estimated. These include airborne bathymetric LiDAR systems~\cite{Marcus_mapping, Hilldale, McKean, DepthLearn}, multispectral imagery~\cite{Misra, Wang}, satellite-derived bathymetry (SDB) using Google Earth engine~\cite{Google}, measurement of water surface elevation~\cite{Inference, Yoon}, measurement of surface velocity through GPS drifters~\cite{Emery} or particle image velocimetry with digital video camera~\cite{Muste}, and thermal imagery~\cite{Puleo, Lima}. In this work, we estimate the bathymetry using surface flow velocity measurements, since they are sensitive to river depth and can be acquired through a number of remote sensing techniques. Once we obtain a reliable measurement of the velocities, we can use one of the several inverse modeling techniques \cite{hojat2020,Wilson,Landon,Lee_Hojat}, to estimate the bathymetries. Here, we have used the principal component geostatistical approach (PCGA)~\cite{PCGA_Lee, PCGA_Kitanidis}. PCGA is a scalable variational inverse modeling method that is accelerated by a low‐rank representation of the covariance matrix; it has been successfully benchmarked for the estimation of bathymetry posterior distributions with superior performance compared to widely used ensemble-based approaches~\cite{Lee_Hojat}.  

The second stage requires a fast solver of the SWEs, given the posterior distribution of the bathymetry as well as the known BCs. Here, we have used machine learning (ML)-based techniques as the fast solvers (see below). These techniques have two major steps, the training step (offline stage) and the prediction step (online stage). The training step typically requires a large amount of data that can be expensive to obtain. However, once the network is trained, the prediction phase can provide accurate and fast results, making these methods suitable for online applications that require rapid update of system state. In this work, we have used the two-dimensional (2D) shallow water module of the U.S\@. Army Corps of Engineers' Adaptive Hydraulics (AdH) model~\cite{AdH} to solve the SWEs numerically (and thus generate the training set data for the offline stage). AdH provides a stabilized finite element (FE) approximation of depths and flow velocities on unstructured 2D meshes and assumes that the bathymetry and BCs are given as inputs to the simulation. For the online stage, we have explored a number of different data-driven ML techniques in our work, which we explain in further detail below.

Data-driven ML techniques have become very popular in many fields such as image processing and natural language processing. This has led to significant interest in leveraging the strength of ML for hydrology applications~\cite{Poggio, review_Sit}, such as flood prediction~\cite{runoff}, water level estimation~\cite{Jeong}, land cover classification~\cite{Abdi}, water quality monitoring~\cite{water_quality}, and water resources management~\cite{Karimi}. In the topic of bathymetry estimation, examples include the use of neural networks (NNs) for detecting the non-linear relationship between reflectance from different spectral bands and water depths~\cite{Ceyhun} and similarly from Landsat images~\cite{Caspian}, bathymetric inversion~\cite{Collins, PCA_DNN_Hojat}, increasing bathymetry resolution using deep-learning-based image super-resolution \cite{superres}, applying ML algorithms for lake bathymetry using satellite images~\cite{Moses, Yunus}, and use of neuro-fuzzy approaches for Quickbird images~\cite{Neuro_fuzzy, Nile}. Some of these works have addressed the bathymetry estimation problem from indirect observations (the first stage of our two-stage process). However, they do not discuss the relationship between the estimated bathymetries and flow velocity prediction with variable BCs.

Since bathymetry and flow velocities are 2D images, the SWEs problem can become computationally expensive for resolutions needed for engineering applications (e.g., on the order of several 10s of thousands of nodes in the computational domain for a kilometer-scale reach). This has led to significant interest in using surrogate models that replace the computationally expensive numerical solver of the SWEs with a fast solver. In particular, reduced-order models (ROMs) attempt to replace the high-fidelity model by fast, dimension-reduced surrogates at the cost of controlled loss of accuracy~\cite{Willcox}. Proper orthogonal decomposition (POD) is one of such methods~\cite{POD} that has been used previously to describe dynamics of SWEs~\cite{POD_SWE}. POD works via approximating the dynamics of a high-dimensional system of equations by finding its reduced-order orthogonal basis via minimization of the projection error. Dynamic mode decomposition (DMD) is another class of ROMs, whose decomposition is based on the dynamics of modes, unlike POD whose decomposition is based on their energy content~\cite{DMD}. POD-NN (POD-neural network) is another class of ROMs that works by first, finding the reduced-order basis via POD, and then using NNs to approximate the coefficients of the reduced-order basis. This method has been used to model steady-states Navier-Stokes equation with variable BCs~\cite{Hesthaven_Ubbiali_18}.

ROMs applications for the SWEs have typically focused on capturing the dynamics of the system by processing the information contained in the snapshots of the time-dependent solutions. However, most have not addressed the influence of the parameters of the PDE, such as the BCs and the bathymetry, on the solution (depths and flow velocities). The POD-NN study in \cite{Hesthaven_Ubbiali_18} does explore the influence of these parameters on the solution for a different set of applications. However, it is limited to linear dimension reduction techniques. Furthermore, the PDE parameters (BC or geometry) considered in POD-NN~\cite{Hesthaven_Ubbiali_18} are low dimensional while the parameter in SWEs problem considered here (the bathymetry) is high-dimensional. To address these issues, we consider several deep learning techniques, equipped with both linear and non-linear dimension reduction that are capable of solving SWEs with variable BCs, (in this case the free surface elevation and the discharge), and spatially varying bathymetry discretized with approximately 40,000 triangular elements and 60,000 degrees of freedom. 

\begin{figure}[htbp]
%\vspace{-0.5cm}
\centering
\includegraphics[width=0.60\linewidth]{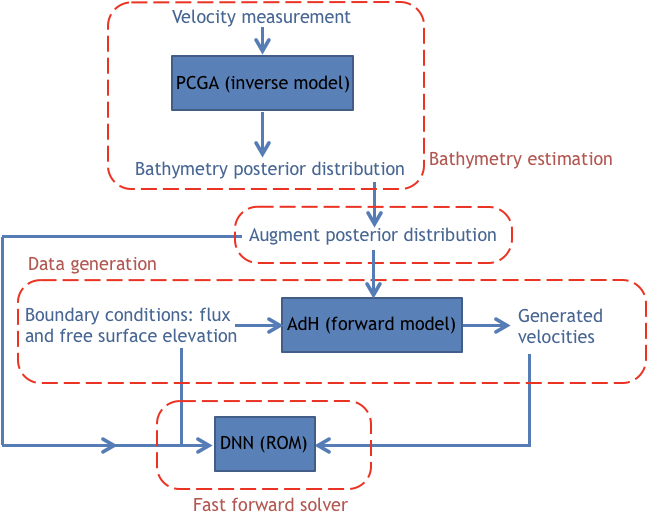}
%\vspace{-1cm}
\caption{The schematic of the development of the forward solver. First, we estimate the posterior distribution of the bathymetry via the PCGA, then augment this distribution to a more general distribution and use AdH to generate velocities. Finally, the bathymetries, BCs, and velocities are fed to DNNs which will be used as fast forward solvers.}
\label{sketch}
\end{figure}

To be more concrete, \cref{sketch} shows the steps in the proposed approach schematically. In many cases, the river bathymetry is estimated using velocity measurements during a short-term data collection campaign with a fixed bathymetry assumption. If we generate the training dataset from the quantified posterior distribution, however, our solver will have less generalization performance when bathymetry changes over time, for example due to sediment deposition or erosion. In order to allow our solvers to include a larger class of bathymetries into their prediction capability, we ``augment'' the posterior distribution via additional sampling process based on typical river topography (see \cref{data_gen}), before feeding them as inputs to the DNN. This allows the solver to be accurate in flow velocity prediction when the updated bathymetry measurements, likely different from the posterior estimate, are provided to it as new inputs. 

The remainder of the paper is as follows. In \cref{meth}, we provide a brief overview of the data-driven approaches being used in our work as fast, forward solvers. In \cref{data_prep}, we discuss the process being used to generate the data, such as bathymetries, BCs, and flow velocities that will be provided to DNNs. In \cref{result_global}, we provide the results of applying different forward solvers as global solvers (predictor of the flow velocity of the whole riverine domain) to a reach of the Savannah river, GA. In \cref{result_local}, we provide the results of applying fast forward solvers as local solvers (predictor of the flow velocity of small segments in the riverine domain) to the Savannah river domain. Finally, in \cref{conclusion} we discuss the major findings from the work and consider potential future directions.

\section{Methods}
\label{meth}

In this work, we use three different deep learning methods to construct fast SWE solvers. These three methods, shortly, are referred to as PCA-DNN (principal components analysis-deep neural network), SE (supervised encoder), and SVE (supervised variational encoder). In this section, we explain these three methods briefly and in \cref{result_global} and \cref{result_local} we show the result of prediction of flow velocity magnitudes for the same dataset using the proposed methods. Flow velocity prediction for different components is also provided in the \href{run:./SI.pdf}{Supplementary Information} file.

\subsection{PCA-DNN}
\label{pca-dnn}

The PCA-DNN method consists of first, a low-rank approximation of data via PCA-based linear projection, and then applying DNN to the reduced-dimension data~\cite{Hesthaven_Ubbiali_18, PCA_DNN_Hojat}. Since a single datapoint of either flow velocity or bathymetry is a high-dimensional 2D image (for instance approximately 20,000 degrees of freedom per variable for the Savannah river example below), training a DNN directly from the original space of bathymetry and BCs to flow velocity requires a very large network, which is computationally restrictive. Furthermore, such a large network would require a large dataset to be trained adequately without being prone to overfitting. Due to these reasons, a dimension reduction step is essential in order to ensure that training of the DNN is manageable.

The PCA step consists of applying the singular value decomposition (SVD) to bathymetries and flow velocities and calculating the low-rank approximation of these data. The SVD method attempts to find the singular vectors and singular values of the data in descending order of singular values. This is equivalent to finding successive linear projections of data for which the reconstruction error of the original data, after projection onto the low-dimensional space and augmentation to original space, is minimal. Low-rank approximation of the data (velocity or bathymetry) can be computed using
\begin{equation} \label{pca_eq}
x_L = U_L x ,
\end{equation}
where $x$ is the $M\times 1$-dimensional vector of any of the datapoint in the original space (velocity or bathymetry), $U_L$ is a $L\times M$-dimensional matrix of the optimal linear projection coefficients, obtained via SVD, and $x_L$ is the $L\times 1$-dimensional low-rank approximation of the original datapoint. In this notation, the rows of the matrix $U_L$ contain transformation coefficients applied to the original data. 

We can use a relationship similar to \cref{pca_eq} to obtain the representation of the low-rank data in the original space. In particular, we can use  
\begin{equation} \label{pca_eq2}
\hat{x} = U_L^\text{T} x_L ,
\end{equation}
in which $\hat{x}$ is the representation of the low-rank datapoint in the original high-dimensional space. In the PCA stage of the PCA-DNN method we use \cref{pca_eq} in order to obtain a low-rank representation of flow velocities and bathymetries. Then in the DNN stage, a DNN is applied to the reduced-dimension data. We use fully connected (FC) layers~\cite{Bishop} in the PCA-DNN structure to map from inputs to velocity outputs. 

\begin{figure}[htbp]
%\vspace{-1cm}
%\hspace{-1cm}
\centering
\begin{subfigure}{.49\textwidth}
\centering
\includegraphics[width=1.05\linewidth]{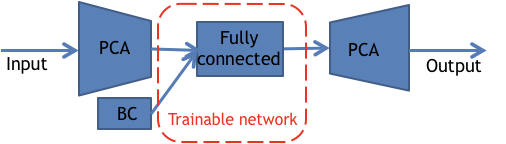}
\caption{PCA-DNN}
\label{pca-dnn_sketch}
\end{subfigure}
\begin{subfigure}{.49\textwidth}
\centering
\includegraphics[width=1.05\linewidth]{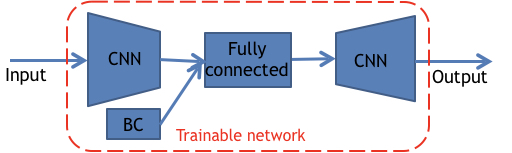}
\caption{SE}
\label{AE_sketch2}
\end{subfigure}
\begin{subfigure}{1.0\textwidth}
\centering
\includegraphics[width=0.70\linewidth]{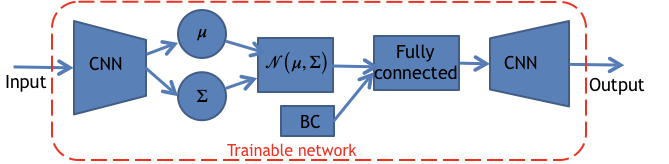}
\caption{SVE}
\label{VAE_sketch}
\end{subfigure}
\caption{(a) Schematic of the PCA-DNN method. In this approach, first, dimension of inputs and outputs are reduced linearly via PCA, and then the low dimensional data are fed to a DNN. (b) Schematic of the SE architecture. Unlike the PCA-DNN, the dimension reduction process in this method is a part of the network architecture and performed in a non-linear fashion. (c) Schematic of the SVE architecture. The difference between SVE and SE is in their bottleneck layer. In SVE, this layer generates a distribution while in SE it is deterministic.}
\label{DNNs_sketch}
\end{figure}

\Cref{pca-dnn_sketch} shows the PCA-DNN method, described in this section, schematically. The ``input" in this figure is the bathymetry, which is fed into the ``fully connected" network after finding its low-rank approximation, obtained from \cref{pca_eq}; the ``BC" is the boundary condition, here taken to be the free-surface elevation and discharge, and the ``output" is the flow velocity, obtained via augmenting their equivalent low-rank values to their original spaces using \cref{pca_eq2}. Note that during the training stage of the PCA-DNN algorithm the low-rank approximations of velocities are fed to the DNN as the output of the network, obtained via applying \cref{pca_eq} to the high-dimensional flow velocity training data. During the prediction stage of the PCA-DNN, \cref{pca_eq2} is used instead on the output of the DNN in order to estimate the velocities (the ``output" in the figure). More details of PCA-DNN method can be found in \cite{PCA_DNN_Hojat} where this approach has been used for prediction of riverine bathymetry from velocity observations.

\subsection{Supervised encoder}
\label{AE}

Due to the close connection between supervised encoders (SEs) and Autoencoders (AEs), we first provide a brief overview of the AE structure and its purpose in deep learning problems. Autoencoders (AEs) are a class of DNNs that are primarily used for non-linear dimension reduction in unsupervised learning~\cite{Kramer, Hinton}. In AE architectures, a high-dimensional input is fed as the input to the network, in the middle of the network (bottleneck) its dimension is reduced, and finally it returns to its original dimension. The network, in its unsupervised fashion, learns a map from a dataset to itself, the so-called auto-associative NNs. However, the strength of the AE is in its ability to simultaneously learn a map from the input to the bottleneck layer, referred to as the encoder, as well as a map from the bottleneck layer to the output, the decoder. The encoder in the AE architecture can be thought of as a non-linear equivalent of \cref{pca_eq} while the decoder is similar to \cref{pca_eq2}. In other words, AE is a non-linear version of the PCA. 

While AE has been primarily introduced as a non-linear dimension reduction technique in unsupervised learning, inspired by its structure, we introduce a supervised encoder (SE) whose input and output, instead of being the same data, are a set of labeled datasets (such as bathymetry and flow velocity). Such architectures are very useful for problems in which the dimension of input and output are much greater than the size of the dataset being used, since using a dense FC architecture requires a very large network which is computationally very expensive to train. Furthermore, such dense networks will lead to severe overfitting without a suitably large dataset.

In this work, we use convolutional layers in the SE structure~\cite{Bishop}; this architecture is referred to as the convolutional SE. Since in a convolutional network, filters are being applied to the whole input image (such as the 2D image of a riverbed profile), it reduces the size of the network significantly by taking advantage of the 2D nature of the input and homogeneity of the extracted features throughout the image. \Cref{AE_sketch2} shows the sketch of the SE used in our work. The ``input" in this figure is the bathymetry, the ``output" is the flow velocity, ``CNN" is the convolutional neural network, and ``fully connected" is the fully connected network.

\subsection{Supervised variational encoder}

We also consider a supervised equivalent of a variational autoencoder (VAE), referred to as SVE (supervised variational encoder), as a forward solver in this work. SVE is a variation of SE in which its bottleneck layer consists of two parallel layers that provide the mean and variance of a multivariate Gaussian distribution, from which the variable that is fed into the decoder (the latent space variable) is sampled. This probabilistic structure imposes a strong regularization effect on the SVE architecture, which potentially leads to better generalization of the network~\cite{VAE_ref}. The SVE has a similar encoder-decoder structure as the one shown in \cref{AE_sketch2}, except the middle layer which is designed as a random number generator, for example, from a multivariate normal distribution. \Cref{VAE_sketch} shows a sketch of the SVE structure. $\mu$ in this figure is the generated mean vector, $\Sigma$ is the generated variance, and ${\cal N}(\mu, \Sigma)$ is the Gaussian distribution from which the latent variable $z$ is generated ($z\sim {\cal N}(\mu,\Sigma)$).

\section{Data preparation}
\label{data_prep}

In this section, we discuss the process being used to obtain the training data for our DNNs (forward solvers). In \cref{bath_est}, we briefly discuss the approach being used to estimate the bathymetry of the river. In \cref{data_gen}, we discuss the generation of the synthetic data, including bathymetries, BCs, and flow velocities that are fed into different DNN architectures. 

\subsection{Bathymetry estimation}
\label{bath_est}

\begin{figure}[htbp]
%\vspace{-1cm}
%\hspace{-1cm}
\centering
\begin{subfigure}{.49\textwidth}
\centering
\includegraphics[width=1.05\linewidth]{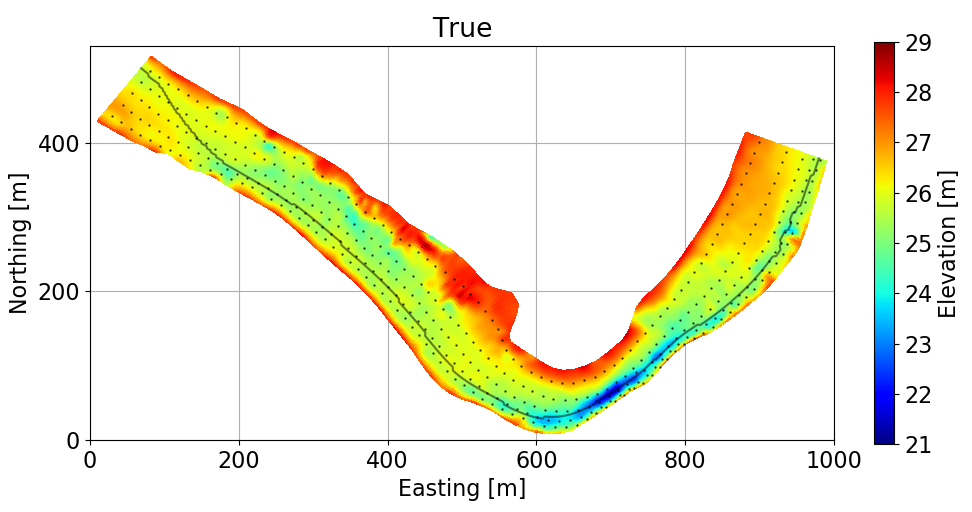}
\caption{}
\label{vel_loc}
\end{subfigure}
\begin{subfigure}{.49\textwidth}
\centering
\includegraphics[width=0.95\linewidth]{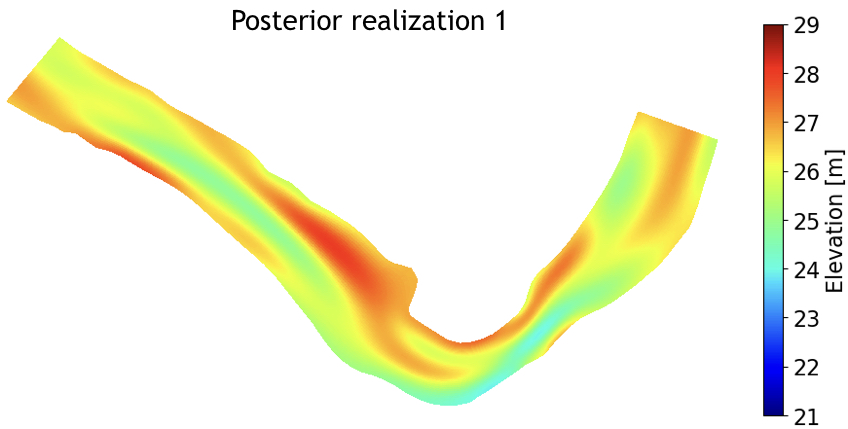}
\caption{}
\label{pcga_example}
\end{subfigure}
%\vspace{-2cm}
\caption{(a) Reference (true) bathymetry of the Savannah river. The black line shows location of the thalweg, the deepest point along the river, at a given cross section. The dots are the 408 measurement locations used as inputs to the PCGA. (b) An example of the generated bathymetry from the PCGA posterior distribution.}
\label{ref_pcga}
\end{figure}

The first step in the data preparation process is applying PCGA to flow velocity observations taken from the river in order to obtain an estimation of the bathymetry in the area of interest. In the following, we refer to this as the PCGA posterior distribution. Here, we have applied PCGA to the roughly one mile reach of the Savannah river near Augusta, GA. The flow velocity measurements in this section are generated synthetically with AdH by first calculating the flow velocities corresponding to the reference bathymetry of the Savannah river ---shown in \cref{vel_loc}---and then applying Gaussian noise with a standard deviation equal to 10$\%$ of the largest simulated flow velocity, in order to ensure the synthetically generated flow velocities include the noise commonly observed in the field observations. Note in this case, the reference bathymetry was obtained from a in-situ survey by Army Corps of Engineers~\cite{Lee_Hojat}.

After we generate the noisy velocities, we provide them as inputs to PCGA. We have input the velocity measurements at only 408 locations (the dots in \cref{vel_loc} collected from five drifter deployment~\cite{Lee_Hojat}), while the AdH simulations and the bathymetry shown in \cref{vel_loc} consist of $41\times 501= 20,541$ mesh nodes (501 nodes in the along-channel and 41 nodes in across-channel direction). This smaller number of measurements is to ensure consistency with the sparsity in common real-world conditions. The approximate nominal spacing between dense nodes is 2.4 m in each direction. The flow velocities predicted by AdH, in here and in what follows, are obtained by simulating the flow over one day to reach quasi-steady state condition.

Once the noisy synthetic velocity measurements are generated, we can use PCGA~\cite{PCGA_Lee} to obtain an estimation of the bathymetry. The estimation is in the form of a distribution (the posterior distribution) that can be represented via its posterior mean riverbed profile and covariance (i.e., Cram\`er–Rao bound). Here we use $n_{PC}=100$ principal components for PCGA. More detail of this number and the PCGA itself can be found in~\cite{PCGA_Lee}. \Cref{pcga_example} shows an example of the bathymetry generated from the posterior distribution obtained from the PCGA inversion. Note that since in general we do not have access to high-resolution bathymetries such as the one shown in \cref{vel_loc}, our information of the bathymetry is primarily based on the PCGA estimation. Therefore, in the remainder of the paper, as was explained in \cref{intro}, we will rely on sampling the posterior distribution shown in \cref{pcga_example} rather than the high-resolution measurements shown in \cref{vel_loc}. The PCGA inversion process took 1.5 hours on a workstation equipped with 48 core Intel(R) Xeon(R) Platinum 8160 @ 2.1 GHz with 128 GB RAM.

\subsection{Data augmentation}
\label{data_gen}
\begin{figure}[htbp]
%\hspace{-1cm}
\centering
\begin{subfigure}{.49\textwidth}
\centering
\includegraphics[width=0.85\linewidth]{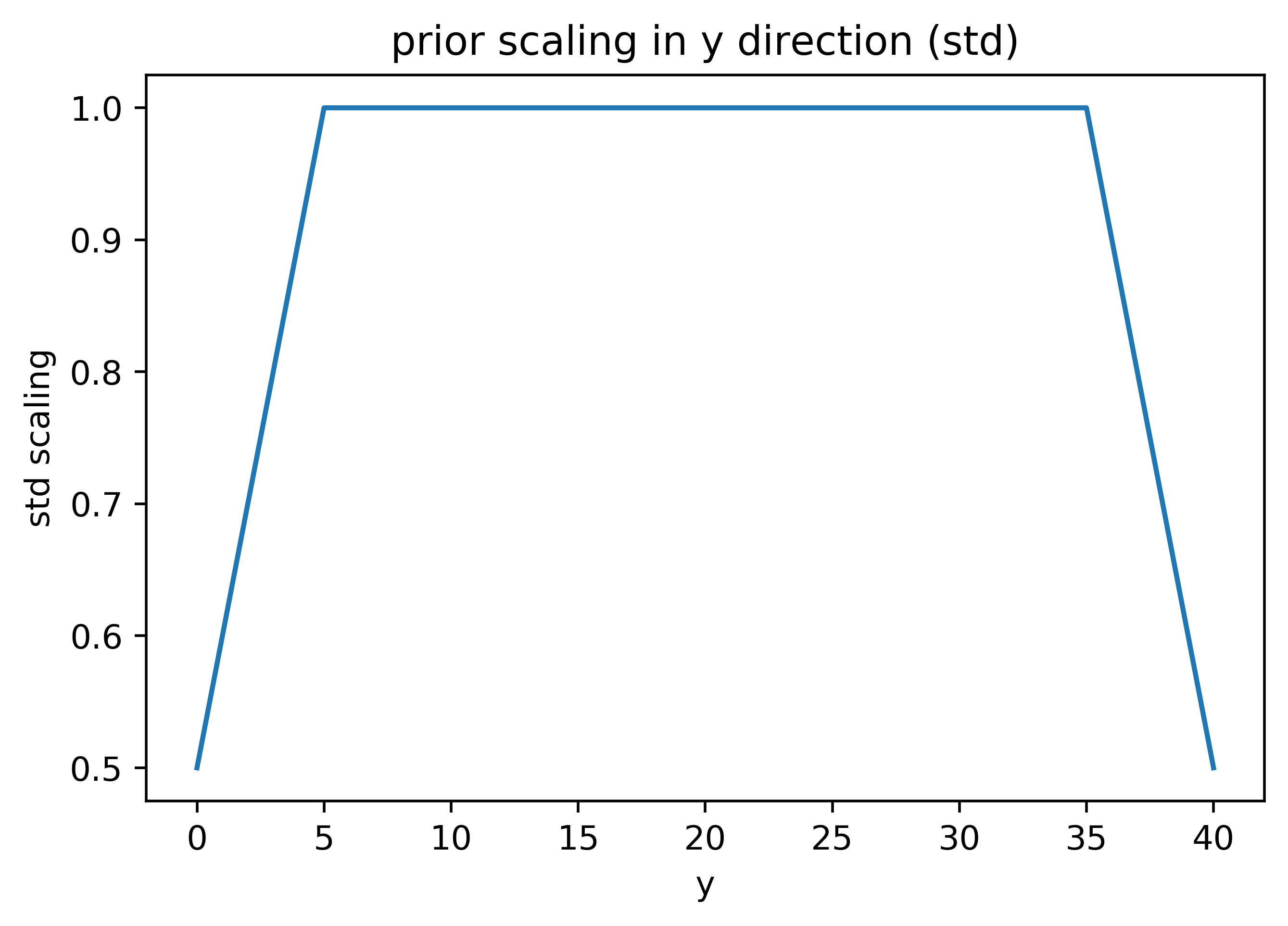}
\vspace{-0.0cm}
\caption{}
\label{y_scale}
\end{subfigure}
\begin{subfigure}{.49\textwidth}
\centering
\includegraphics[width=1.0\linewidth]{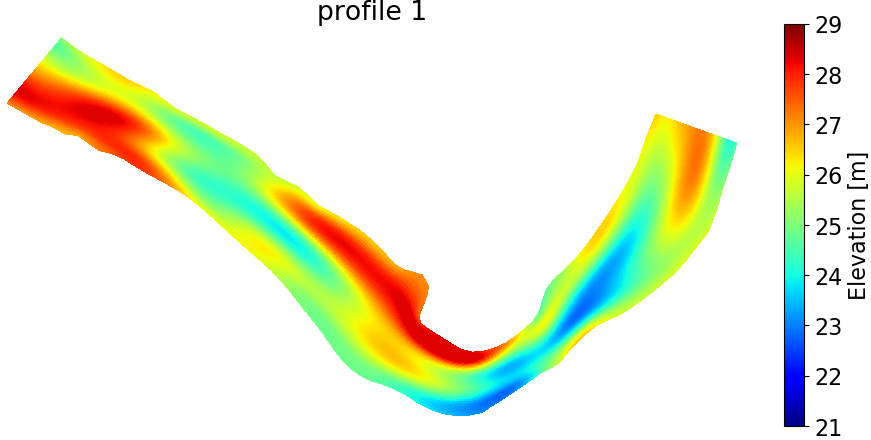}
\vspace{-0.0cm}
\caption{}
\label{prof_example}
\end{subfigure}
%\vspace{-2cm}
\caption{(a) The weighting factor applied to the standard deviation of the generated bathymetries in the across-river direction. (b) An examples of the generated bathymetry after adding Gaussian kernel and weighting factor to the PCGA posterior distribution. This profile corresponds to the same profile shown in \cref{pcga_example} after the augmentation.}
\label{y_prof_ex}
\end{figure}

While the PCGA posterior distribution provides a reasonable estimate of the uncertainty associated with the currently available dataset, we also consider an additional augmentation of the training data in order to broaden the range of bathymetries for which the proposed forward solvers are valid for (e.g., when the bathymetry changes over time) and further avoid overfitting. To perform the augmentation, the synthetic data that are fed to the DNN architectures are generated by adding a Gaussian kernel of the following form to the PCGA-estimated bathymetry realizations:
\begin{equation}
\text{cov}(x,y)= \beta^2 \exp\left( -\frac{\Delta x^2}{l_x^2}-\frac{\Delta y^2}{l_y^2} \right)
\label{cov_x_y}
\end{equation}
Here, $\beta= 1.2$ m, $l_x= 115$ m, and $l_y= 29$ m ($x$ is the along-river direction while $y$ is the across-river direction; see \cref{vel_loc}). We then use the Kronecker product~\cite{K_delta} representation of separable covariance matrices to assign the variable standard deviations in the across-the-river direction ($y$) with a weighting factor in the form shown in \cref{y_scale}. The role of this function is to capture the fact that the variations of the generated bathymetries near the shore are generally smaller than in the middle of the river. \Cref{prof_example} shows the profiles shown in \cref{pcga_example} after the Gaussian kernel of \cref{cov_x_y} with the weighting factor are applied. Note, the resulting set of training bathymetries is not intended to span a wide range of river types, but rather should better reflect the river in question under a wider range of possible BCs and bathymetric changes.

\begin{figure}[htbp]
\hspace{-2.2cm}
\centering
\begin{subfigure}{.32\textwidth}
\centering
\includegraphics[width=1.40\linewidth]{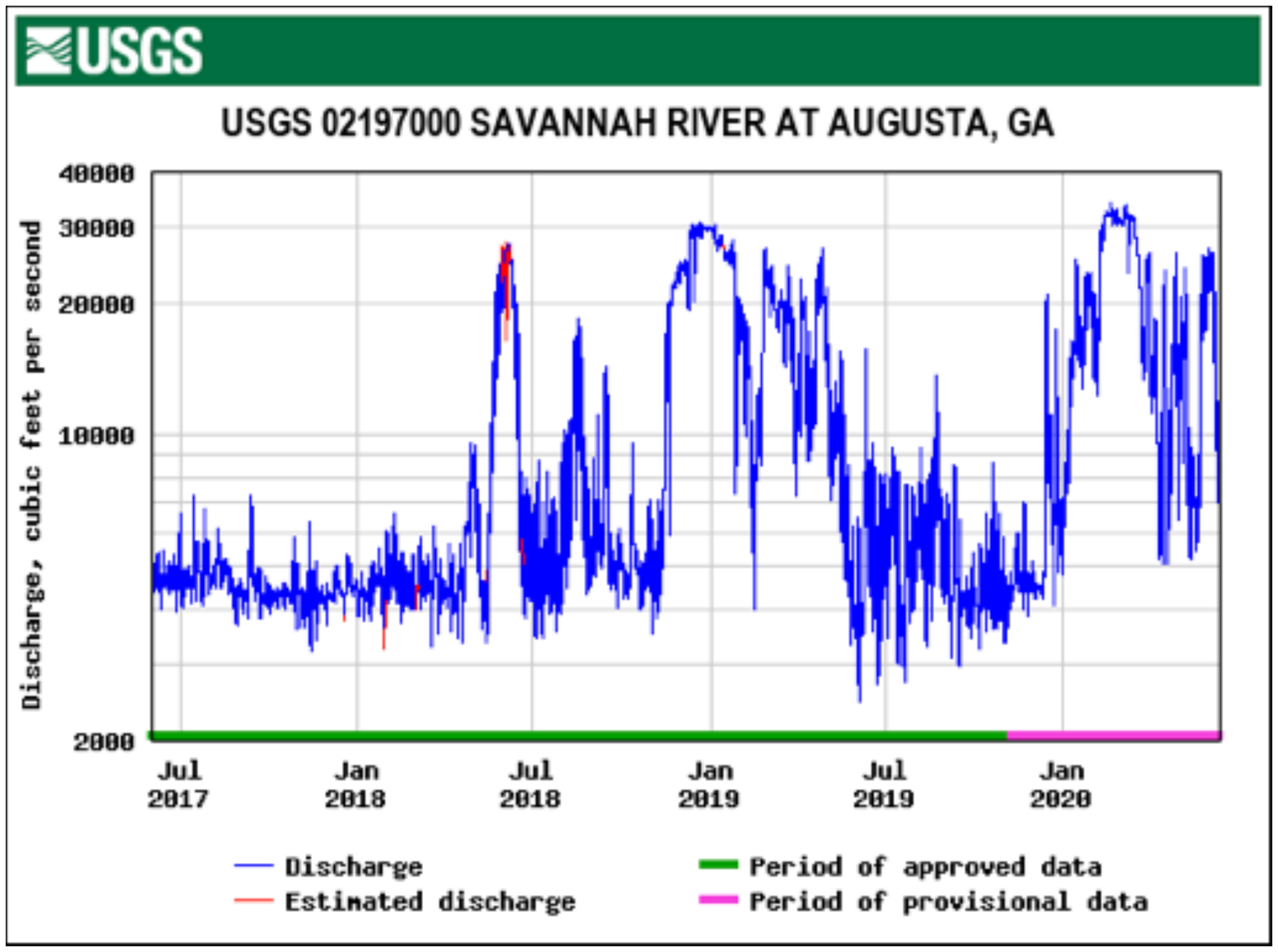}
%\caption{}
\label{Hollandmodel}
\end{subfigure}
\hspace{-0.4cm}
\begin{subfigure}{.32\textwidth}
\centering
\includegraphics[width=1.40\linewidth]{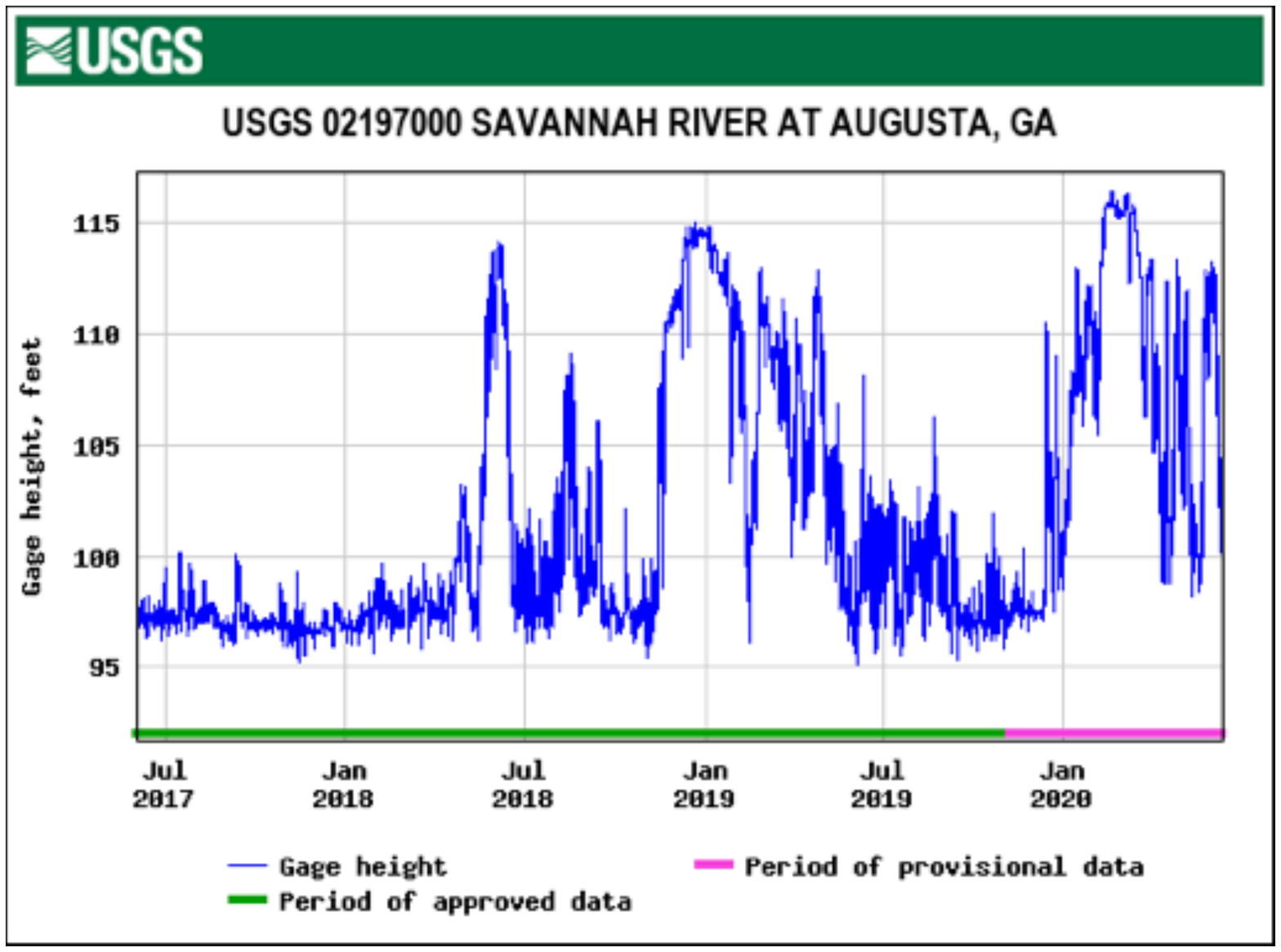}
%\caption{}
\label{ab_initiomodel}
\end{subfigure}
\hspace{0.7cm}
\begin{subfigure}{.33\textwidth}
\centering
\vspace{0.6cm}
\includegraphics[width=1.20\linewidth]{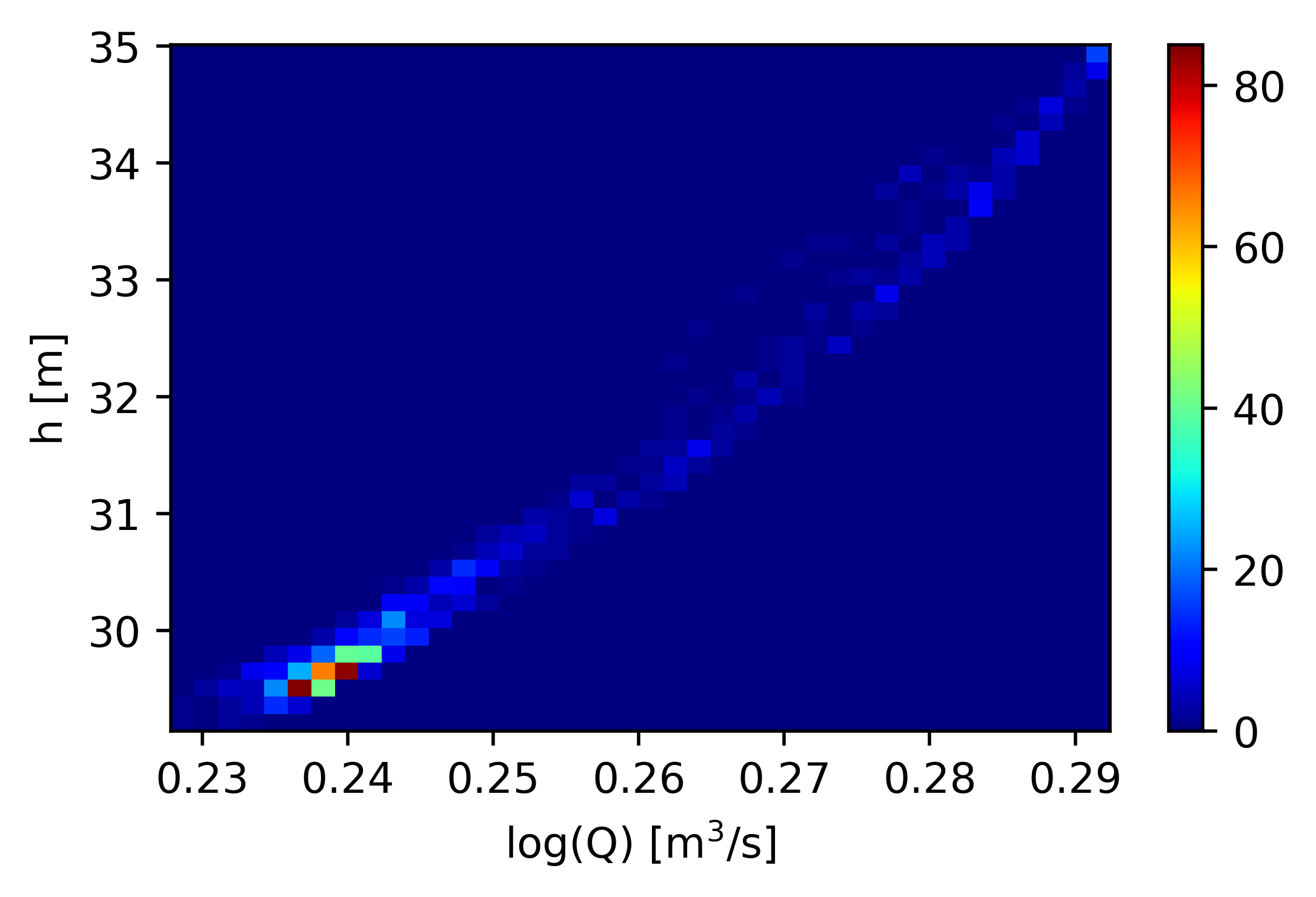}
%\caption{}
\label{ab_initiomodel}
\end{subfigure}
\vspace{-1cm}
\caption{Discharge (left), free-surface elevation (middle), and the joint distribution of free-surface elevation and discharge (right) of the Savannah river over a three-year period obtained from the USGS.}
\label{BC_USGS}
\end{figure}

After we generate the synthetic bathymetries, such as the one shown in \cref{prof_example}, we generate BCs. In order to be consistent with the actual BCs observed over time, we have extracted the BCs from the United States Geological Survey (USGS) gage data taken over a three-year period~\cite{USGS_web}. The two BCs (discharge and free-surface elevation) are presented in \cref{BC_USGS}. The survey includes the BCs from June 2017 to June 2020. We have also plotted the free-surface elevation versus discharge over this period in \cref{BC_USGS}. We observe, as expected, that there is a strong correlation between the two BCs as they change over time.

Once we generate the bathymetries and BCs, we can provide them as inputs to AdH in order to obtain the flow velocities. Here, we have generated 450 bathymetries, in the form presented in \cref{prof_example}, and for each profile we have generated ten different BCs. The BCs are obtained by first generating a uniformly distributed discharge value $Q \in(85,\;840)$ m$^3/$s, and then finding its corresponding free-surface elevation from a parabolic function that is fit to the joint distribution observed in \cref{BC_USGS}. In other words, we assume that there is only one free-surface elevation value associated with any discharge value and these discharge-free surface elevation pairs are located on a parabola that is fitted to their joint distribution (\cref{BC_USGS}). The number of BCs and bathymetries that we have chosen leads to a dataset of size $450\times 10=4500$. Once a synthetic bathymetry as well as a pair of BCs have been generated, we can use AdH to calculate the flow velocities. Finally, these bathymetry/BC/flow velocity datasets are fed to the DNNs to obtain forward solvers, as will be explained in \cref{result_global} and \cref{result_local}.

\section{Global solver}
\label{result_global}

Here, we provide the result of applying different DNN architectures as global solvers for the flow velocity prediction of the Savannah river. The global solver takes the bathymetry of the entire domain of the river and outputs the flow velocities for the entire domain, hence the name ``global solver." The discussion of the local solver, which takes bathymetry of small segments of the river in question and outputs the velocity for that segment, is left to \cref{result_local}. In \cref{global_full}, we discuss the performance of the solvers when full bathymetry information is available. In \cref{global_partial}, the performances when no (or incomplete) bathymetry information is available will be discussed.

\subsection{Performance in the presence of full bathymetry measurement}
\label{global_full}

In \cref{perform_glob}, we show the result of applying different solvers to the Savannah river domain and compare their performances. In \cref{latent}, we discuss the ability of different solvers to find a low-dimensional representation of the SWEs dynamics.

\subsubsection{Performance of different methods}
\label{perform_glob}

\Cref{error_global} summarizes the root mean square errors (RMSEs) in estimating flow velocity magnitudes using different methods. Here, we used 10$\%$ of our data for validation and a different set of unused 450 input-output pairs for testing. We also compare performance of our methods with a purely linear model. The linear model is similar to PCA-DNN, except the DNN part of the architecture has linear activation functions. In order to have a fair comparison between different methods, we used the same latent space dimension in all methods, equal to 50 (see \cref{latent} for further detail regarding this choice). The errors in \cref{error_global} for SVE and SE are significantly lower than PCA-DNN and the linear model, indicating that the non-linear dimension reduction contained in SVE and SE is more accurate than a linear, PCA-based approach. For example, the RMSEs of SE or SVE are on average about 2 cm$/$s lower than PCA-DNN or the linear model, which is significant considering that the errors of PCA-based methods are on the order of 5--6 cm$/$s. The errors in the velocity prediction of northing and easting directions, separately, are provided in the \href{run:./SI.pdf}{Supplementary Information} file.

\begin{table}[htbp]
    \centering
    \begin{tabular}{lllll}
        \toprule
        \multirow{2}{*}{Error} & \multicolumn{4}{c}{Fast forward solver}\\
        \cmidrule{2-4} \cmidrule{5-5}
        {} & PCA-DNN & PCA with linear map & SE & SVE \\
        \midrule
        Train set RMSE [m/s]   & 0.0515 & 0.0514 & {\bf 0.0269} & 0.0286\\
        Validation set RMSE [m/s] & 0.0570 & 0.0571 & {\bf 0.0374} & 0.0398\\
        Test set RMSE [m/s] & 0.0546 & 0.0544 &  {\bf 0.0381} & 0.0398\\
        \bottomrule
    \end{tabular}
    \caption{Comparison between the error of different global solvers when predicting the magnitude of the flow velocity.}
    \label{error_global}
\end{table}

\Cref{hype_global} summarizes the hyperparameters used in different global solvers. The table shows the different parameter values used in our networks during the hyperparameter tuning along with the final chosen value, which had the best performance (shown in bold in the table). All hyperparameter values are the ones used in both easting and northing directions, except the regularization coefficient which had different values for easting and northing directions (as can be seen in the table as well). In the study, we observed that the choice of using or not using batch normalization and the regularization coefficient were more influential than the activation function, learning rate, and the batch size. We used a decay rate of 0.001 in all networks and Adam optimizer with mean squares error as the loss function. We also used gradient descent (GD) and stochastic gradient descent (SGD) optimizers for a number of architectures and observed better performance of Adam in all of them. We also tried different number of neurons in different layers. The training of the networks with the best performances takes about about 5 minutes for PCA-DNN and SE, and about 10 minutes for the SVE, on a GPU cluster with 1x NVIDIA V100 GPU, 1x Intel(R) Xeon(R) CPU @ 2.00 GHz and 25 GB RAM. The SVD step of the PCA-DNN was pre-computed on the 48 core machine where the PCGA inversion was performed and took about one minute.

\begin{table}[htbp]
    \centering
    \begin{tabular}{p{0.27\textwidth}p{0.2\textwidth}p{0.2\textwidth}p{0.2\textwidth}p{0.22\textwidth}}
        \toprule
        \multirow{2}{*}{DNN hyperparameter} & \multicolumn{3}{c}{Fast forward solver}\\
        \cmidrule{2-4} \cmidrule{5-5}
        {} & PCA-DNN & SE & SVE \\
        \midrule
        Type of layers   & Fully connected & Convolutional & Convolutional\\
        Batch normalization   & \{yes, {\bf no}\} & \{yes, {\bf no}\} & \{yes, {\bf no}\}\\
        Number of hidden layers   & \{{\bf 1},2,3,4,5,6\} & \{4,{\bf 6}\} & \{4,{\bf 6}\}\\
        Data normalization  & \{{\bf yes}, no\} & \{{\bf yes}, no\} & \{{\bf yes}, no\}\\
        Act. func. (hidden layer)  & \{{\bf tanh}, ReLU\} & \{{\bf tanh}, ReLU\} & \{{\bf tanh}, ReLU\}\\
        Act. func. (output layer)  & \{{\bf linear}, Sigmoid\} & \{{\bf linear}, Sigmoid\} & \{{\bf linear}, Sigmoid\}\\
        Batch size & \{8,{\bf 32},256,full\} & \{8,{\bf 32},256,full\} & \{8,{\bf 32},256,full\}\\
        Learning rate & \{0.01,{\bf 0.001},$10^{-4}$\} & \{0.01,{\bf 0.001},$10^{-4}$\} & \{0.01,{\bf 0.001},$10^{-4}$\}\\
        Reg. coeff. (easting) & \{0,0.00001,0.0001, 0.001,{\bf 0.01},0.1,1\} & \{0,0.00001,{\bf 0.0001}, 0.001,0.01,0.1,1\} & \{0,0.00001,{\bf 0.0001}, 0.001,0.01,0.1,1\}\\
        Reg. coeff. (northing) & \{0,0.00001,0.0001, 0.001,{\bf 0.01},0.1,1\} & \{0,0.00001,{\bf 0.0001}, 0.001,0.01,0.1,1\} & \{0,0.00001,0.0001, {\bf 0.001},0.01,0.1,1\}\\
        \bottomrule
    \end{tabular}
    \caption{The hyperparameters used in different global solvers. The parameters in bold are the final values used in networks with the best performances. Act. func. is the activation function and reg. coeff. is the regularization coefficient. }
    \label{hype_global}
\end{table}

\begin{figure}[htbp]
\centering
\begin{subfigure}{.49\textwidth}
\centering
\includegraphics[width=1.0\linewidth]{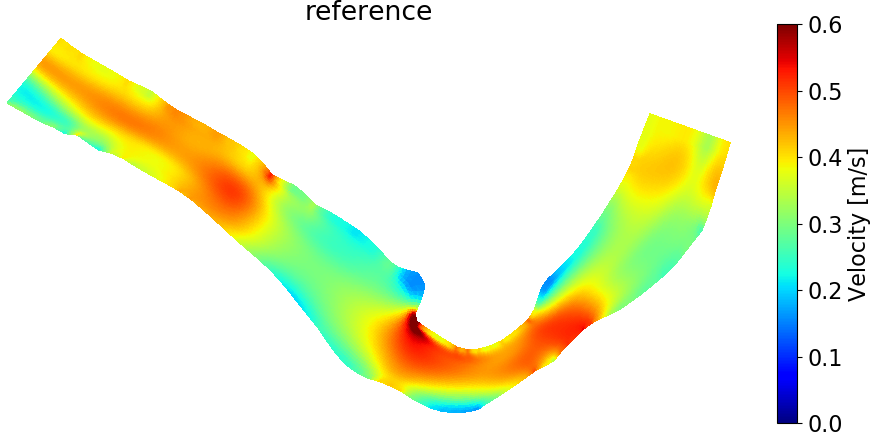}
%\caption{}
\label{ab_initiomodel}
\end{subfigure}
\begin{subfigure}{.49\textwidth}
\centering
\includegraphics[width=1.0\linewidth]{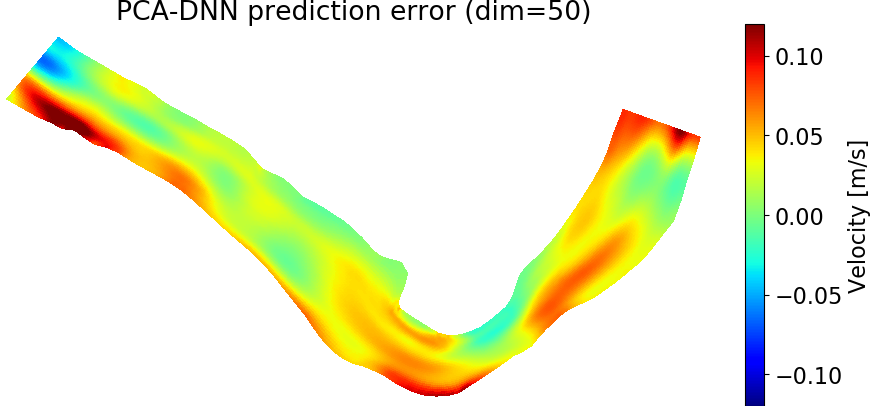}
%\caption{}
\label{Hollandmodel}
\end{subfigure}
\begin{subfigure}{.49\textwidth}
\centering
\includegraphics[width=1.0\linewidth]{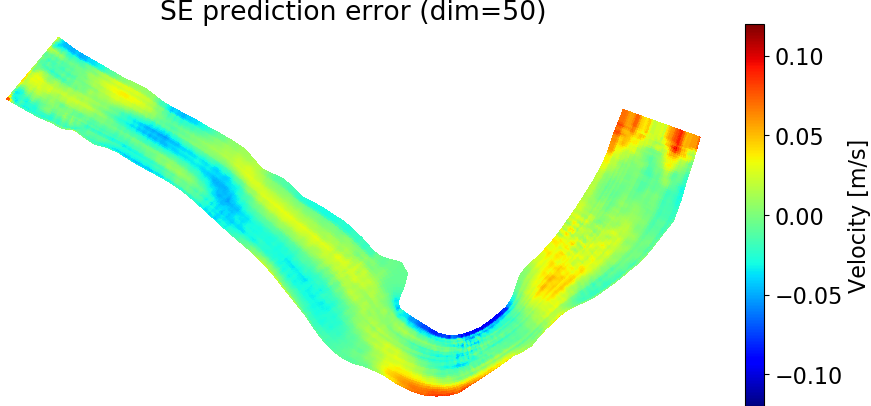}
%\caption{}
\label{ab_initiomodel}
\end{subfigure}
\begin{subfigure}{.49\textwidth}
\centering
\includegraphics[width=1.0\linewidth]{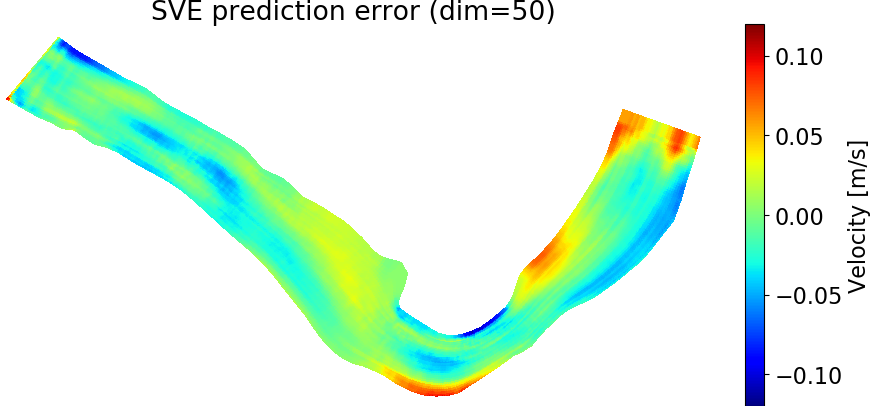}
%\caption{}
\label{ab_initiomodel}
\end{subfigure}
\caption{Examples of the error in the prediction of the velocity magnitudes for different global solvers for small BC values ($z_f= 29.9$ m and $Q= 146.1$ m$^3$/s). SE and SVE outperform PCA-DNN.}
\label{plots_global_low}
\end{figure}

\begin{figure}[htbp]
\centering
\begin{subfigure}{.49\textwidth}
\centering
\includegraphics[width=1.0\linewidth]{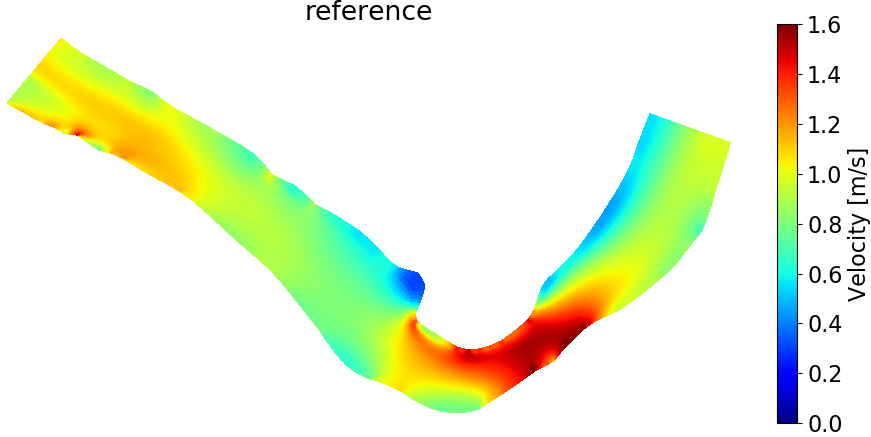}
%\caption{}
\label{ab_initiomodel}
\end{subfigure}
\begin{subfigure}{.49\textwidth}
\centering
\includegraphics[width=1.0\linewidth]{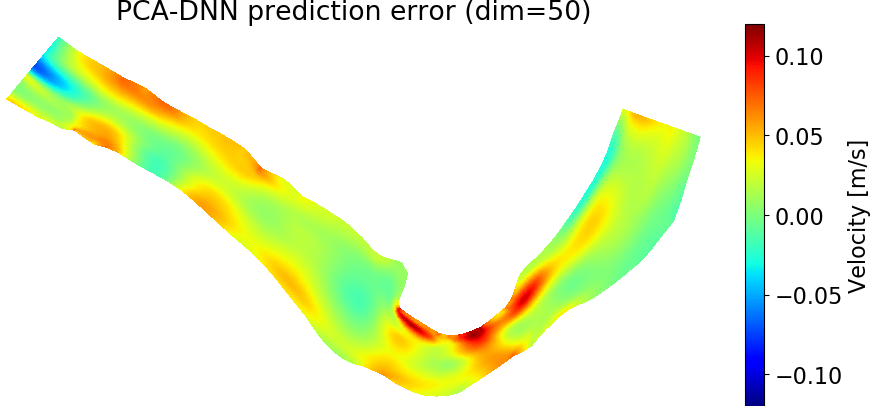}
%\caption{}
\label{Hollandmodel}
\end{subfigure}
\begin{subfigure}{.49\textwidth}
\centering
\includegraphics[width=1.0\linewidth]{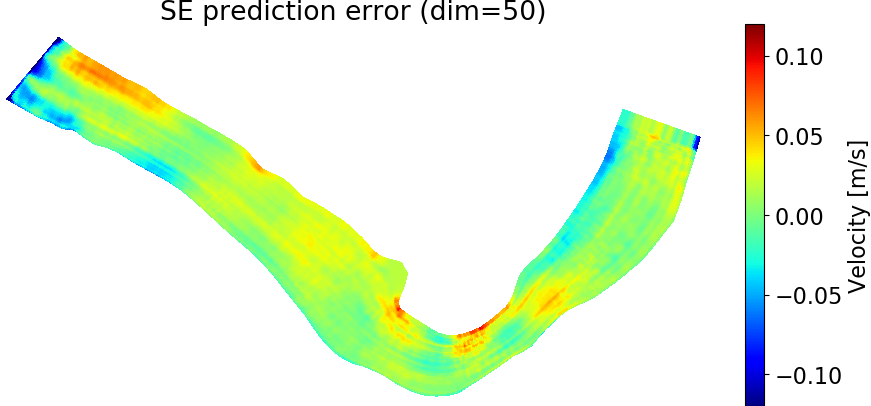}
%\caption{}
\label{ab_initiomodel}
\end{subfigure}
\begin{subfigure}{.49\textwidth}
\centering
\includegraphics[width=1.0\linewidth]{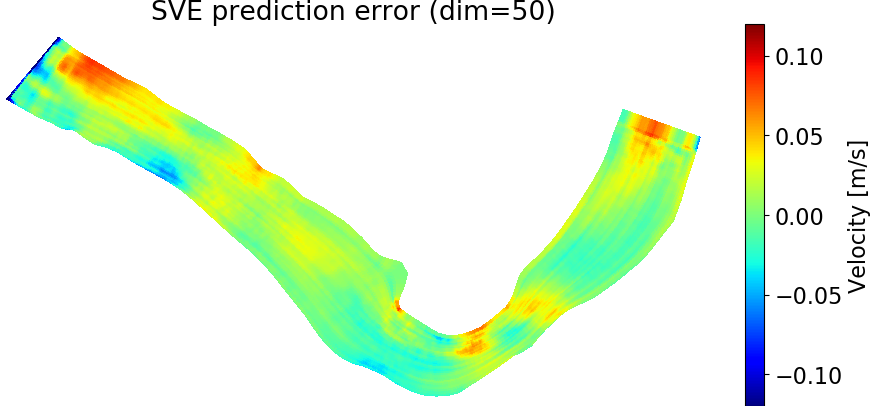}
%\caption{}
\label{ab_initiomodel}
\end{subfigure}
\caption{Examples of the error in the prediction of the velocity magnitudes for different global solvers for higher magnitude BC values ($z_f= 34.8$ m and $Q= 836.6$ m$^3$/s). SE and SVE outperform PCA-DNN.}
\label{plots_global_high}
\end{figure}

\Cref{plots_global_low} compares the performance of different methods when predicting the flow velocity magnitude (see \cref{vel_loc}) of one of the members of the test dataset with small BC values (free-surface elevation $z_f= 29.9$ m and discharge $Q= 146.1$ m$^3$/s). \Cref{plots_global_high} shows a similar comparison for high BC values ($z_f= 34.8$ m and $Q= 836.6$ m$^3$/s). We observe that in both cases SE and SVE perform better than PCA-DNN, consistent with the result of \cref{error_global}. This could be due to the linear dimension reduction technique being used in this approach, which fails to capture non-linear features present in the data with 50 principal components (PCs). This type of behavior is detectable for other datapoints as well. The reference and predicted velocity profiles for the easting and northing directions for the two BCs that are presented in \cref{plots_global_low} and \cref{plots_global_high} can be found in the \href{run:./SI.pdf}{Supplementary Information} file. The predictions of global solvers such as the ones shown in \cref{plots_global_low} or \cref{plots_global_high} take around one second on the GPU workstation where the training was performed.

\begin{figure}[htbp]
\centering
\hspace{-0.45cm}
\begin{subfigure}{.32\textwidth}
\centering
\includegraphics[width=1.1\linewidth]{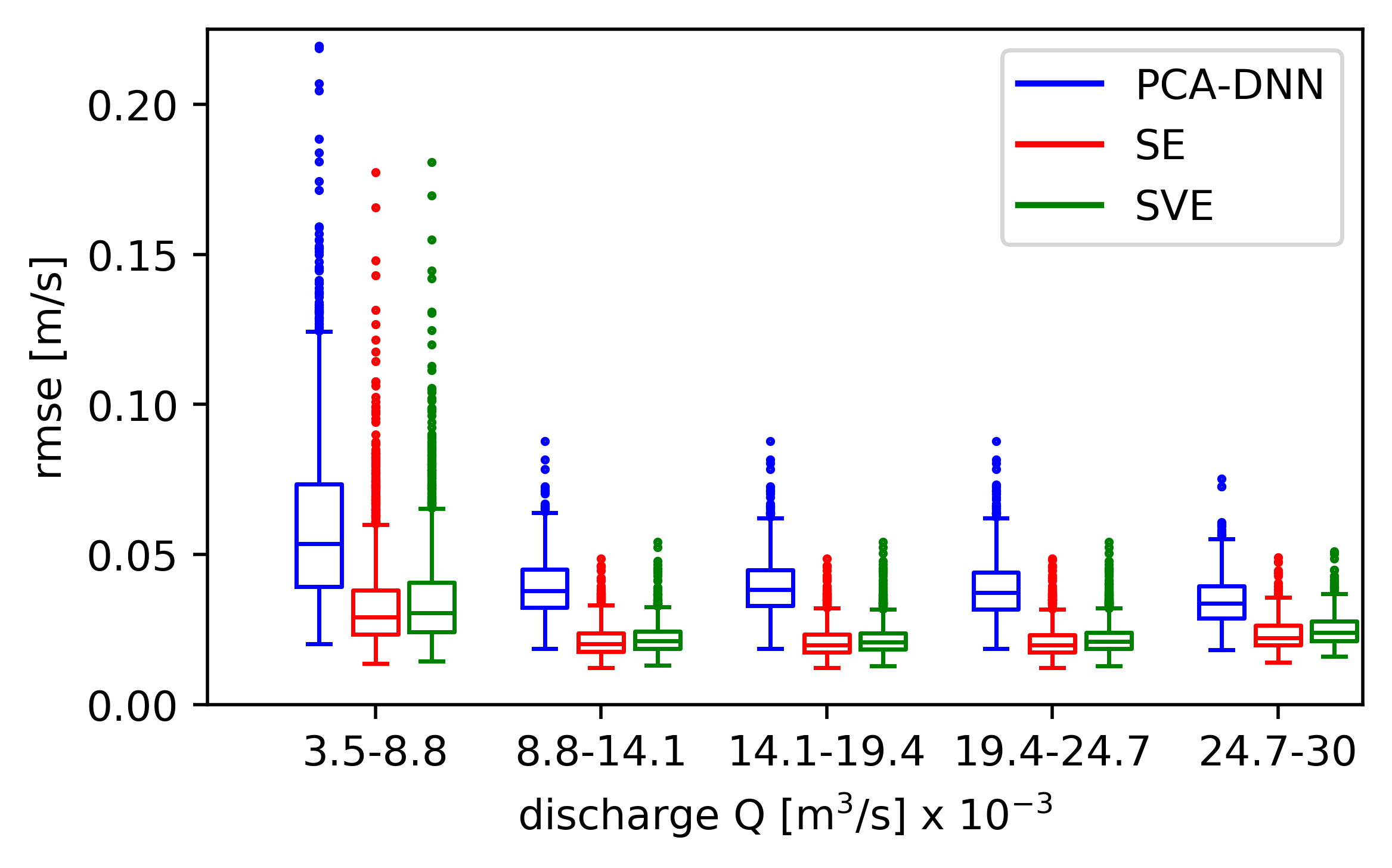}
\caption{}
\label{error_dist}
\end{subfigure}
\hspace{0.2cm}
\begin{subfigure}{.32\textwidth}
\centering
\includegraphics[width=1.05\linewidth]{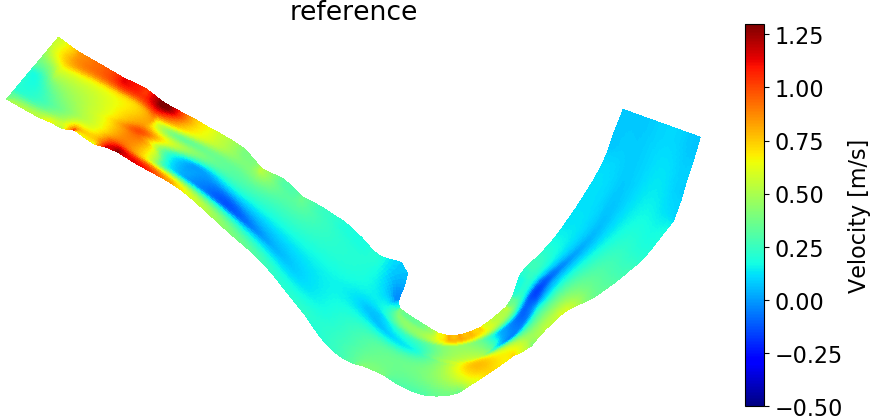}
\vspace{0.2cm}
\caption{}
\label{worst_small}
\end{subfigure}
\hspace{-0.2cm}
\begin{subfigure}{.32\textwidth}
\centering
\includegraphics[width=1.05\linewidth]{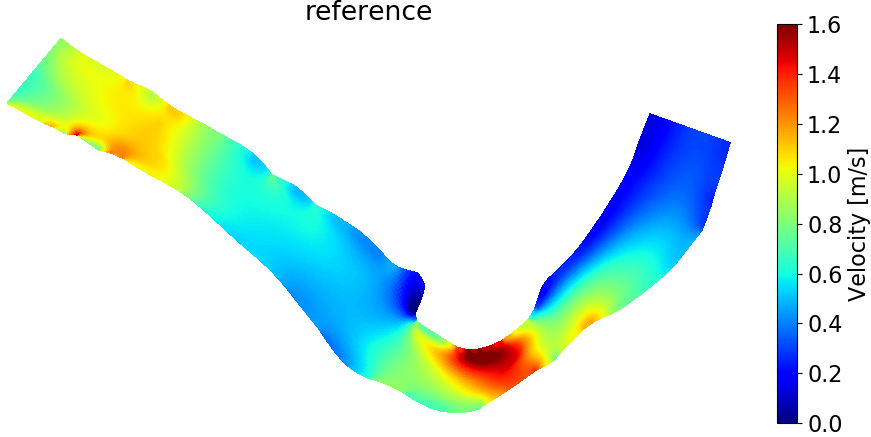}
\vspace{0.2cm}
\caption{}
\label{worst_large}
\end{subfigure}
\caption{(a) Distribution of the error over different discharge values for PCA-DNN, SE, and SVE. (b) Flow velocity in the easting direction for the datapoint with the largest error. (c) Flow velocity for another datapoint with the same bathymetry but larger discharge value. Larger relative variation of the velocities at small BCs is recognizable in the figure.}
\label{dist_worst_case}
\end{figure}

\Cref{error_dist} shows the distribution of errors in the prediction of the magnitude of the flow velocity as a function of the discharge value for different datapoints (train, validation, and test dataset) for different solvers. The figure compares the errors of different methods for 5 different intervals of the same size---very small, small, medium, large, and very large BCs. The boxes show the data in each interval whose errors are between the first ($Q_1$) and third ($Q_3$) quartiles, and the outliers are the data outside the interval $[Q_1-1.5(Q_3-Q_1),Q_3+1.5(Q_3-Q_1)]$~\cite{boxplot}. We observe that for most of the discharge values, the error is distributed uniformly as a function of the discharge value. In particular, for very large discharge values where flow rates and associated hazards like flood risk are high, we do not see any increase in the error, although the velocity values become significantly larger. We also observe an increase in the error for datapoints with very small discharge values. This could be due to the fact that at smaller discharge values, the free-surface elevation is also very small, and consequently, the topography of the riverbed has a stronger influence on the surface flow velocity, leading to a more complicated dynamics. 

\Cref{worst_small} and \cref{worst_large} compare the flow velocity of the easting direction for the datapoint with the largest error (\cref{worst_small}) with another datapoint with the same bathymetry but larger discharge value (\cref{worst_large}). More complicated dynamics at small BCs is recognizable in the figure. For instance, the velocity in the river bend is usually large (see \cref{plots_global_low}, \cref{plots_global_high}, or \cref{worst_large}) while we see a different pattern in \cref{worst_small}. The larger relative variation of velocity can also contribute to larger error for small BCs. For instance, the variation of the velocity in \cref{worst_small} and \cref{worst_large} over the domain is similar, while the velocity value in the case of \cref{worst_small} is smaller. We can solve the issue of large errors at very small-BC datapoints by either assigning a higher weight in the loss function for datapoints with small BCs, or generating more datapoints with small BCs.

\subsubsection{Latent space interpretability}
\label{latent}
In this section, we study the effect of different elements of the latent space on the flow velocity prediction. This provides useful information on the ability of different methods to find linear/non-linear transformation of the data onto the low dimensional latent space that are capable of capturing dynamics of the system (the SWEs). In particular, we perturb different components of the latent space variable for the train, validation, and test datasets and evaluate their influence on the flow velocity prediction by calculating the change in the RMSEs. Our perturbation is of the form of 
\begin{equation}
z^\text{perturbed}_{j,i}= z_{j,i}+ 2\sigma_i
\label{pert}
\end{equation}
in which $z_{j,i}$, for $i \in \{1,\dots,50\}$ and $j \in \{1,\dots,N\}$ is the $i$-th component of the latent space for the $j$-th datapoint in the dataset. $z^\text{perturbed}_{j,i}$ is its value after the perturbation, and $\sigma_i$ is the standard deviation of the $i$-th component calculated over the whole train, validation, or test datasets. 

After performing the perturbation for all components, we can calculate the change in the RMSEs due to perturbation of the $l$-th element of the latent space using
\begin{equation}
\Delta \text{RMSE}_l= \Bigg| \sqrt{\frac{\sum_{j,k} \left( f_{j,k}({\bf z}^{\text{perturbed},l}_{j})-y_{j,k} \right)^2}{N}} - \sqrt{\frac{\sum_{j,k} \left( f_{j,k}({\bf z}_{j})-y_{j,k} \right)^2}{N}} \Bigg| ,
\label{pert_RMSE}
\end{equation}
in which ${\bf z}^{\text{perturbed},l}_{j}$ is a latent space vector for the $j$-th datapoint that has same elements as the original values, except the $l$-th component. That is, 
\begin{eqnarray*}
\left[{\bf z}^{\text{perturbed},l}_{j}\right]_i&=& z_{j,i}, \mbox{ for } i \in \{1,\dots,50\} \mbox{ and } i\neq l, \mbox{ while } \\
\left[{\bf z}^{\text{perturbed},l}_{j}\right]_l&=& z_{j,l}+2\sigma_l
\end{eqnarray*}
Also, ${\bf z}_j=(z_{j,1},\dots,z_{j,50})$, $y_{j,k}$ is the $k$-th component of the $j$-th output (flow velocity) for each dataset member $j$, $N$ is the total number of points at which the error has been calculated ($N=$ dimension of the flow velocity $\times$ number of river profiles), and $f_{j,k}$ is the $k$-th component of the map from the latent space to the flow velocity for the $j$-th dataset member. For the PCA-DNN method, this is a map from the input of the DNN to the flow velocity while for the SE or SVE, a map from the input of decoder to its output. Using \cref{pert_RMSE} we can study the influence of different variables of the latent space on the velocity, separately. This sensitivity analysis can then provide useful information regarding how important any of the components in the low-dimensional representation are in defining the dynamics of the system within each solver as well as different solvers compared to each other. 

\begin{figure}[htbp]
\centering
\begin{subfigure}{.32\textwidth}
\centering
\includegraphics[width=1.025\linewidth]{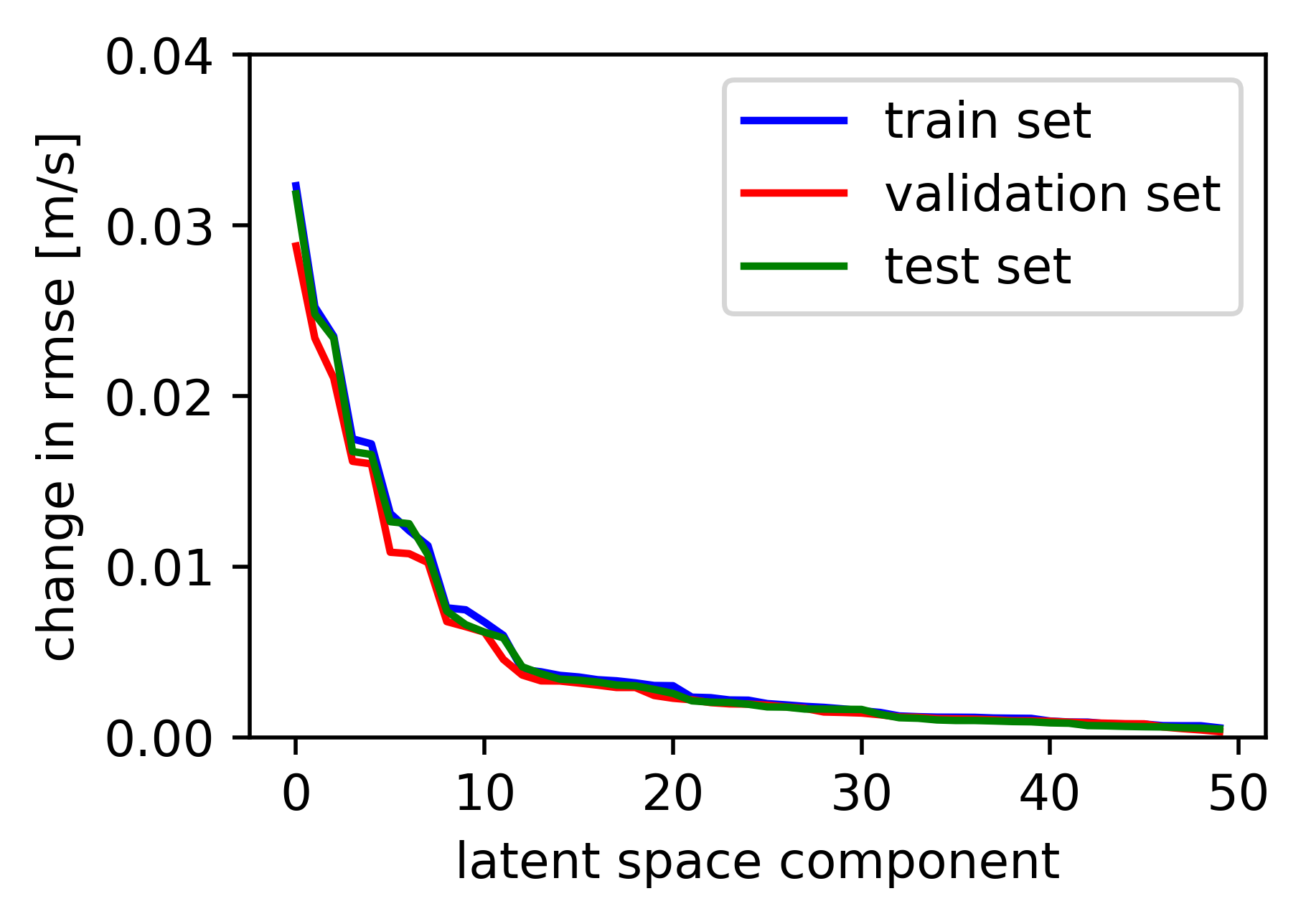}
\caption{PCA-DNN}
\label{ab_initiomodel}
\end{subfigure}
\begin{subfigure}{.32\textwidth}
\centering
\includegraphics[width=1.0\linewidth]{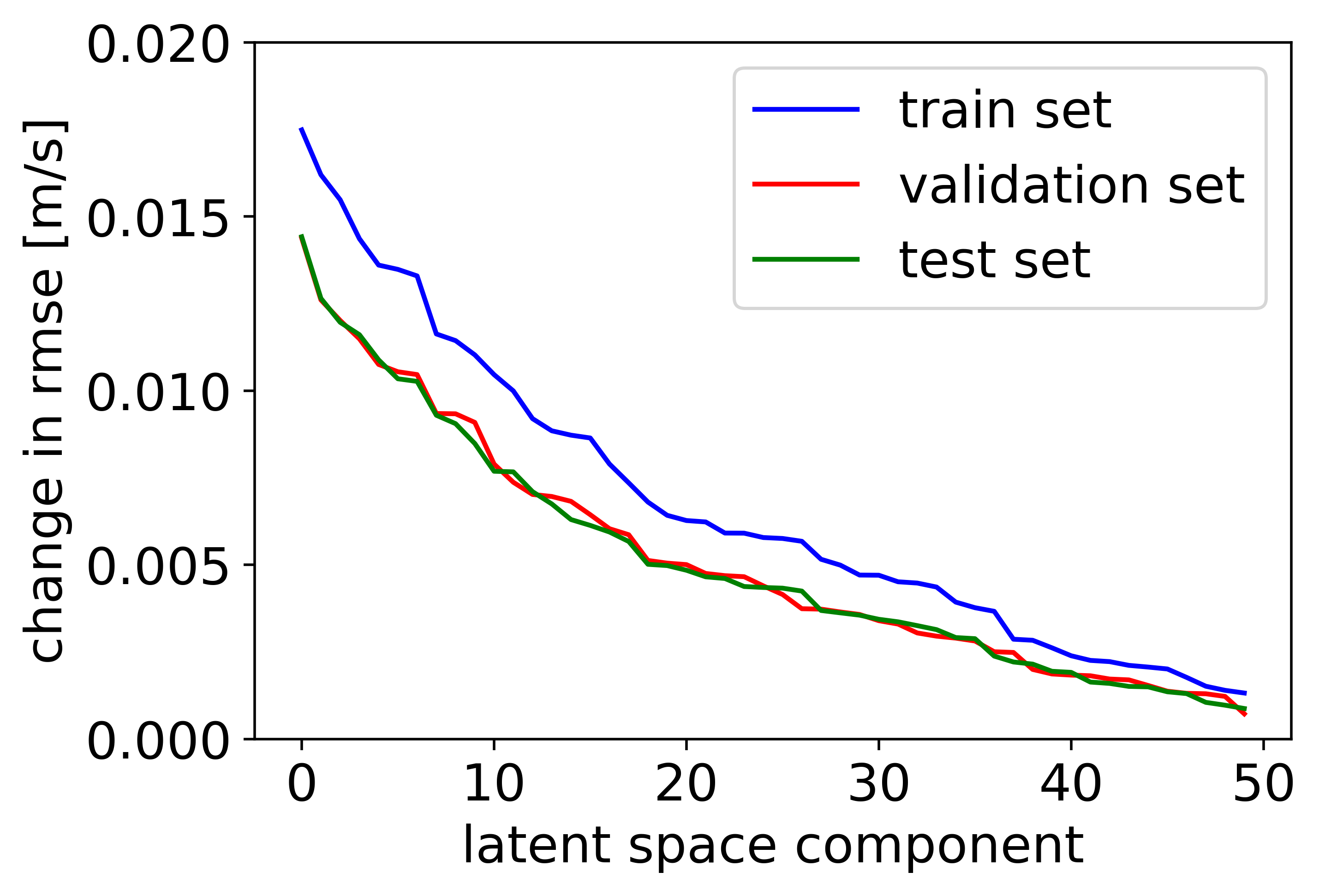}
\caption{SE}
\label{Hollandmodel}
\end{subfigure}
\begin{subfigure}{.32\textwidth}
\centering
\includegraphics[width=1.05\linewidth]{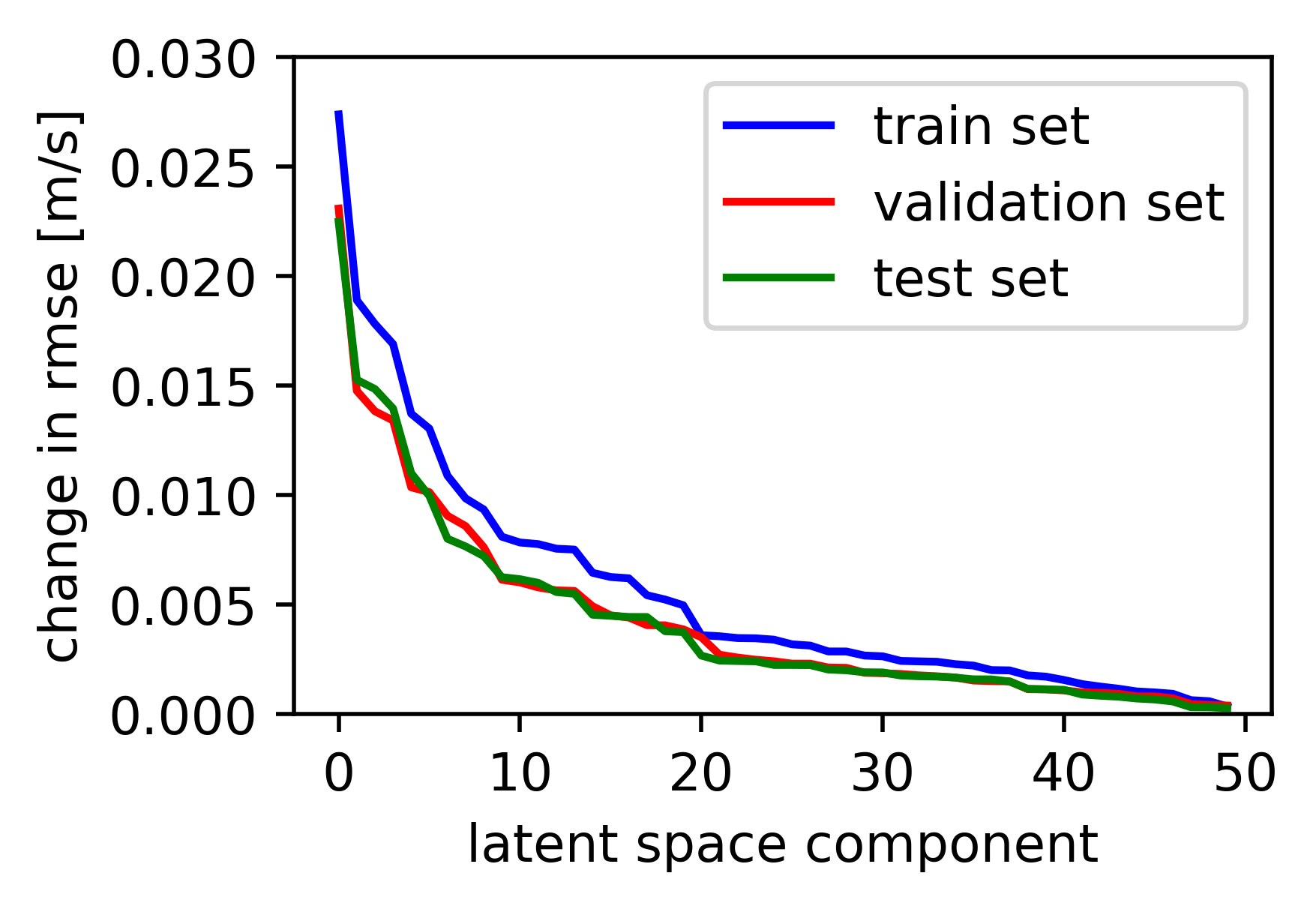}
\caption{SVE}
\label{ab_initiomodel}
\end{subfigure}
\caption{Influence of different latent space elements on the output of different global solvers for the train, validation, and test datasets. The convergence to zero as the component index increases indicates that the chosen dimension of the latent space is sufficient to reach the desired accuracy.}
\label{plots_errs}
\end{figure}

\Cref{plots_errs} shows the change in RMSEs caused by perturbing different latent space components of the easting direction solvers for train, validation, and test datasets. We observe that in all approaches the change in RMSE reaches a near-zero value, implying that the dimension of latent space chosen here, 50, is sufficient to capture the dynamics of the system. 

\begin{figure}[htbp]
\centering
\hspace{-1.2cm}
\begin{subfigure}{.325\textwidth}
\centering
\includegraphics[width=1.2\linewidth]{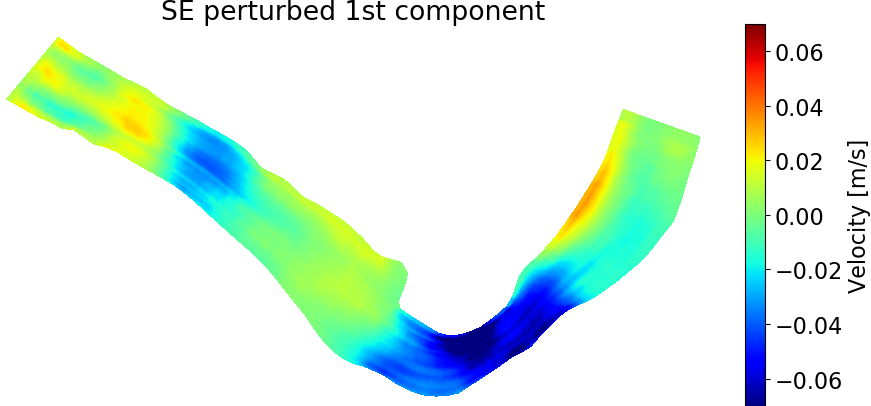}
%\caption{}
\label{ab_initiomodel}
\end{subfigure}
\hspace{-0.50cm}
\begin{subfigure}{.325\textwidth}
\centering
\includegraphics[width=1.2\linewidth]{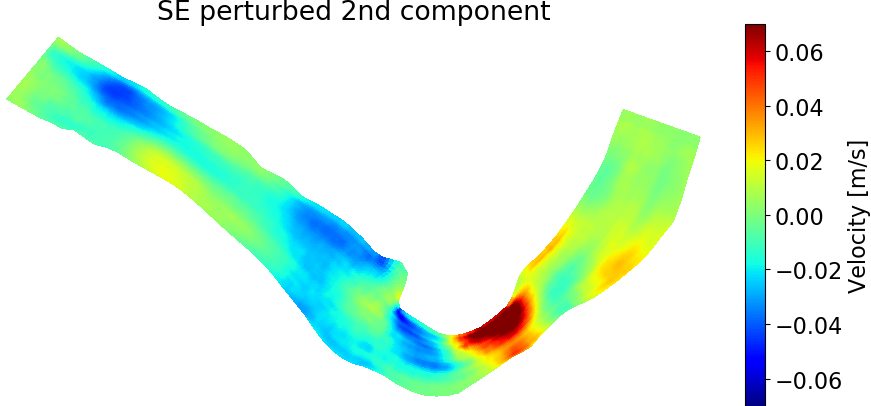}
%\caption{}
\label{Hollandmodel}
\end{subfigure}
\hspace{-0.50cm}
\begin{subfigure}{.325\textwidth}
\centering
\includegraphics[width=1.2\linewidth]{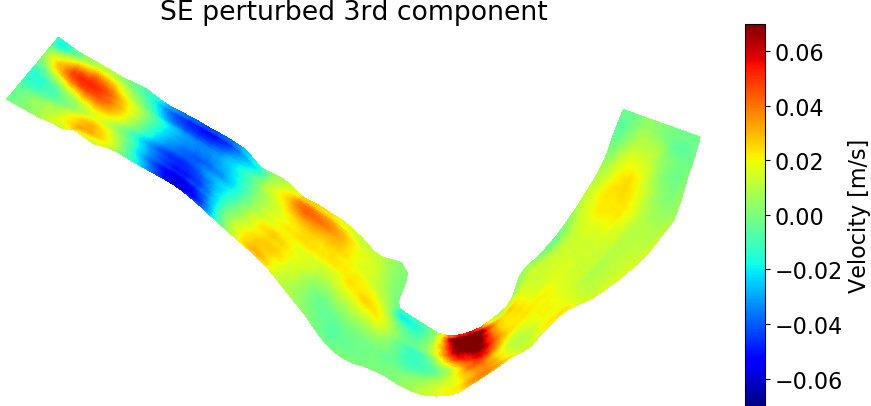}
%\caption{}
\label{ab_initiomodel}
\end{subfigure}
\caption{An example of the influence of different latent space elements on the output of the SE. These plots are based on the perturbation of \cref{pert}.}
\label{errs_AE_component}
\end{figure}

\Cref{errs_AE_component} shows the difference between the perturbed (via \cref{pert}) and the original predicted profile for the same datapoint that the results of \cref{plots_global_low} were based on for the SE method in easting direction. We observe that in all three components there is a larger variation in the river bend, consistent with the larger velocities observed in \cref{plots_global_low} and \cref{plots_global_high} in this region. Furthermore, the oscillatory nature of these modes are consistent with common wavelet or Fourier mode decompositions, implying a physically meaningful interpretation of the dynamics of the system (see fig. 14 of \cite{Lee_Hojat} for an example of the eigenmodes of the prior covariance of the PCGA).

\subsection{Performance in the presence of uncertain bathymetry}
\label{global_partial}

In \cref{no_bathy}, we show the result of using our solvers for the prediction of the flow velocities when no additional bathymetry measurement other than the PCGA posterior distribution is available. In \cref{partial_bathy}, we show the results when bathymetry at a limited number of cross sections is available.

\subsubsection{Performance with indirect velocity observation}
\label{no_bathy}

The results presented in \cref{perform_glob} provide informative evaluation metrics of different algorithms as forward solvers, that is, flow velocity predictors provided with bathymetry and BCs assuming the reference (true) bathymetries are known completely. In practice, however, there are many situations in which we do not have access to direct measurement of bathymetries, and all (or most) of our information about the bathymetry must come from the solution of an inverse problem with an associated level of uncertainty. In order to evaluate the performance of the reduced-order DNN solvers under these conditions, we next move to a series of experiments based on the posterior distribution for bathymetry obtained by PCGA estimate outlined in Step 1 above.

\begin{figure}[htbp]
\centering
\begin{subfigure}{.49\textwidth}
\centering
\includegraphics[width=1.0\linewidth]{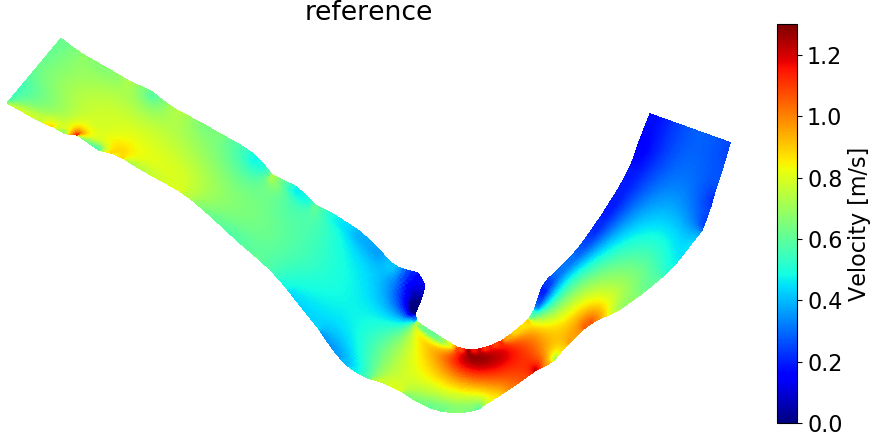}
%\caption{}
\label{ab_initiomodel}
\end{subfigure}
\begin{subfigure}{.49\textwidth}
\centering
\includegraphics[width=1.0\linewidth]{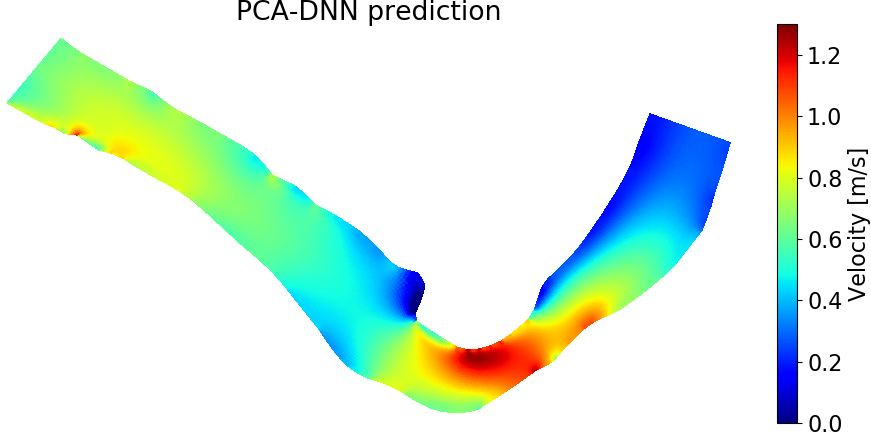}
%\caption{}
\label{Hollandmodel}
\end{subfigure}
\begin{subfigure}{.49\textwidth}
\centering
\includegraphics[width=1.0\linewidth]{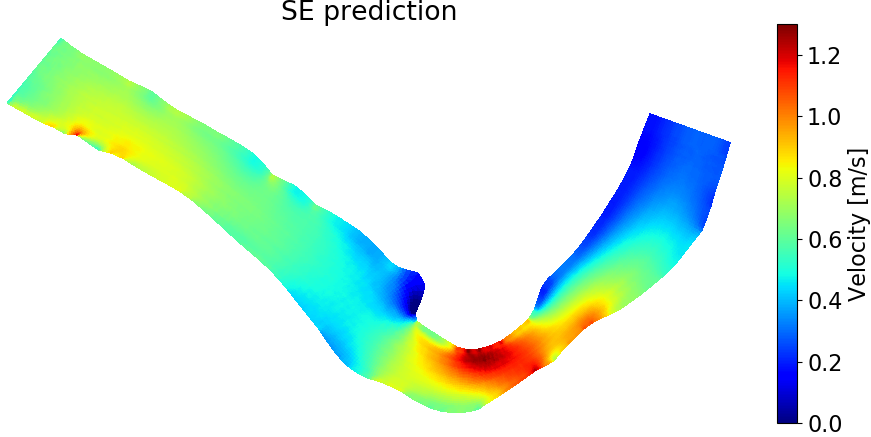}
%\caption{}
\label{ab_initiomodel}
\end{subfigure}
\begin{subfigure}{.49\textwidth}
\centering
\includegraphics[width=1.0\linewidth]{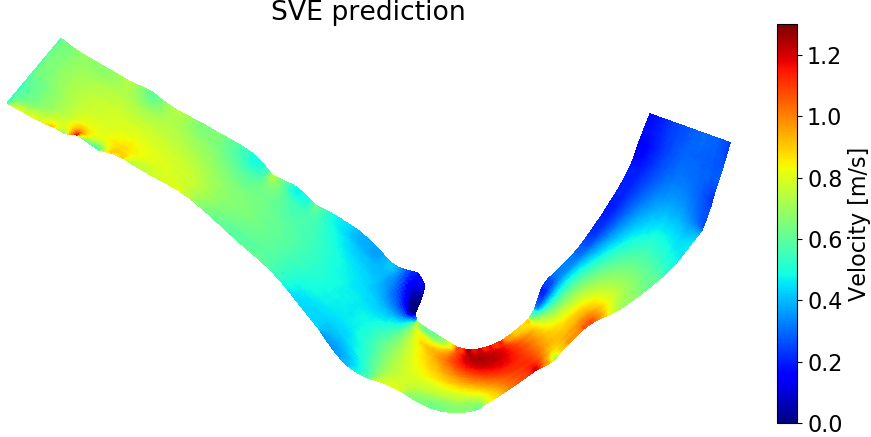}
%\caption{}
\label{ab_initiomodel}
\end{subfigure}
\caption{Predicted mean velocities for different global solvers at $z_f= 33.9$ m and $Q= 651.2$ m$^3$/s. The ``reference" corresponds to the AdH prediction of the mean of the flow velocity when bathymetries are generated from the PCGA posterior distribution.}
\label{plots_global_post}
\end{figure}

\begin{figure}[htbp]
\centering
\begin{subfigure}{.49\textwidth}
\centering
\includegraphics[width=1.0\linewidth]{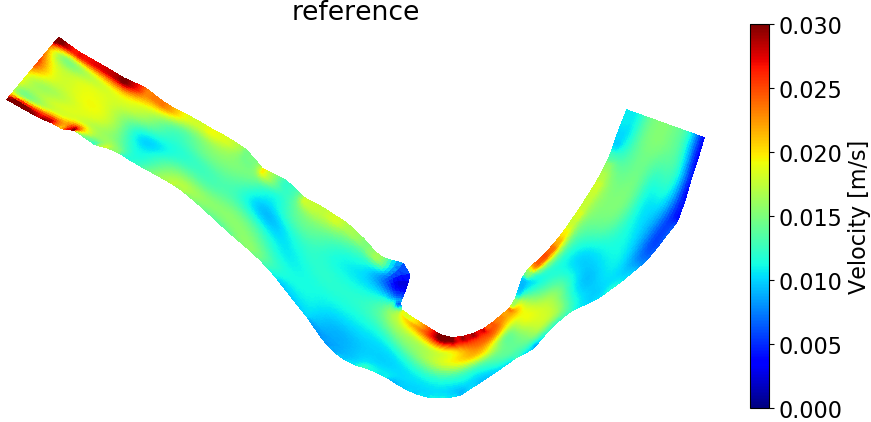}
%\caption{}
\label{ab_initiomodel}
\end{subfigure}
\begin{subfigure}{.49\textwidth}
\centering
\includegraphics[width=1.0\linewidth]{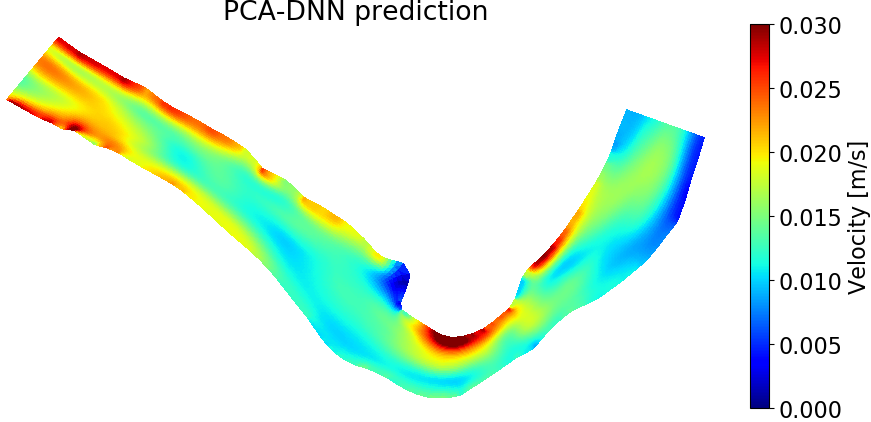}
%\caption{}
\label{Hollandmodel}
\end{subfigure}
\begin{subfigure}{.49\textwidth}
\centering
\includegraphics[width=1.0\linewidth]{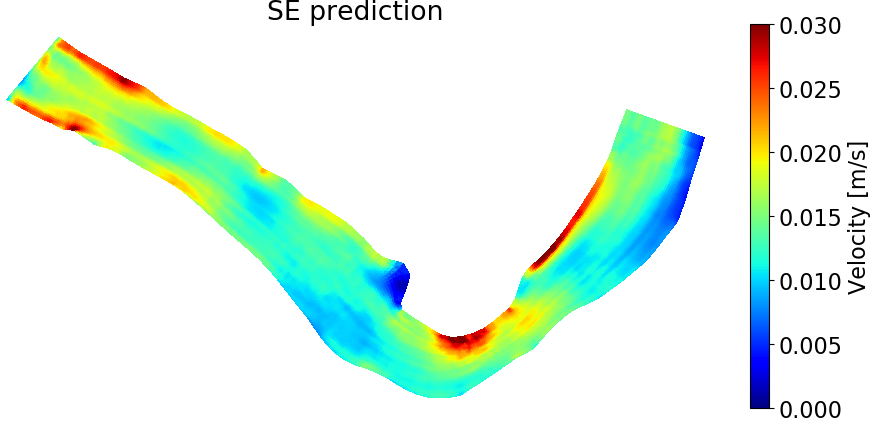}
%\caption{}
\label{ab_initiomodel}
\end{subfigure}
\begin{subfigure}{.49\textwidth}
\centering
\includegraphics[width=1.0\linewidth]{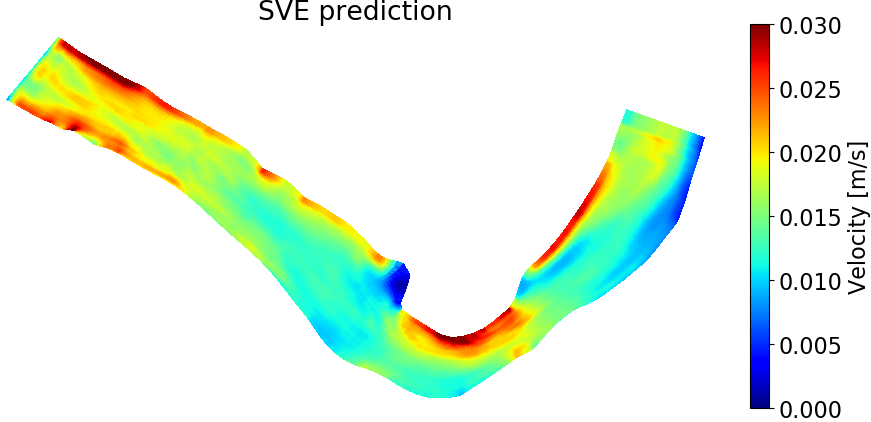}
%\caption{}
\label{ab_initiomodel}
\end{subfigure}
\caption{Predicted standard deviation of velocities for different global solvers at $z_f= 33.9$ m and $Q= 651.2$ m$^3$/s. The ``reference" corresponds to the AdH prediction of the standard deviation of the flow velocity when bathymetries are generated from the PCGA posterior distribution.}
\label{plots_global_post_std}
\end{figure}

\Cref{plots_global_post} and \cref{plots_global_post_std} show the reference mean and standard deviation of flow velocities in the easting direction obtained from AdH as well as the predicted mean and standard deviations obtained from the different reduced-order solvers, respectively. The BCs for the simulations are  $z_f= 33.9$ m and $Q= 651.2$ m$^3$/s. The results are based on first, generating bathymetries directly from the PCGA posterior distribution, and then providing these profiles as inputs to either the AdH or any of the DNNs (with the given BCs); finally, the mean and standard deviation of their predicted velocities are calculated and plotted in \cref{plots_global_post} and \cref{plots_global_post_std}. These plots are based on generating 100 profiles from the bathymetry posterior distribution. 

We observe that in both figures, all methods have been successful in finding the mean and the uncertainty. The small differences observed in \cref{plots_global_post_std} is due to the fact that the standard deviation of the estimated bathymetry via the PCGA is quite small and consequently its predicted velocity has a standard deviation which is difficult for any of the algorithms to capture accurately. Note the small values observed in \cref{plots_global_post_std} (0--0.03 m/s) compared to velocity magnitudes observed in \cref{plots_global_low} or \cref{plots_global_high} (0--1.25 m/s). The high accuracy of the results provided in \cref{plots_global_post} and \cref{plots_global_post_std} implies that even when indirect observations are available, we can use our same global solvers, which are trained on the estimated bathymetry distribution from PCGA with augmentation, to predict the distribution of flow velocities as the BCs change. 

\subsubsection{Performance with additional incomplete bathymetry measurements}
\label{partial_bathy}
\begin{figure}[htbp]
\centering
\begin{subfigure}{.49\textwidth}
\centering
\vspace{0.3cm}
\includegraphics[width=1.02\linewidth]{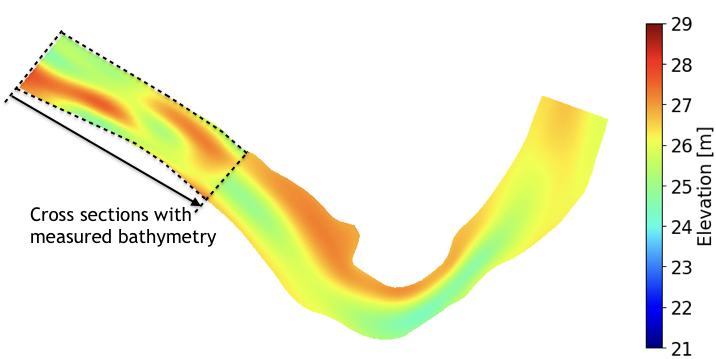}
\vspace{0.01cm}
\caption{}
\label{loc_section}
\end{subfigure}
%\hspace{0.1cm}
\begin{subfigure}{.49\textwidth}
\centering
%\vspace{-1cm}
\includegraphics[width=0.95\linewidth]{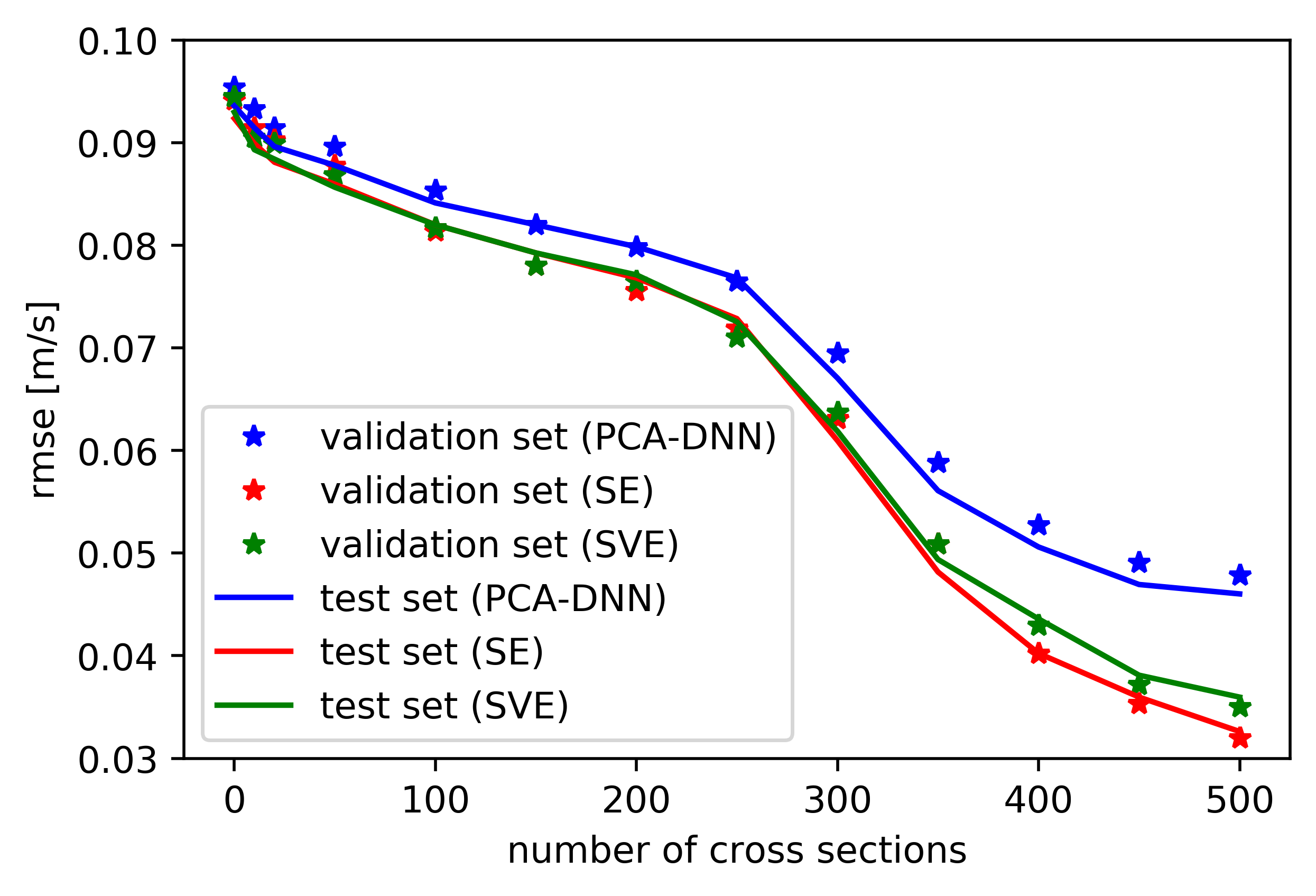}
\vspace{-0.3cm}
\caption{}
\label{err_cross}
\end{subfigure}
\caption{(a) The locations of the (first 150) cross sections with measured bathymetry. (b) The change in the error of the flow velocity prediction for different global solvers as a function of the number of cross sections used for the bathymetry measurement.}
\label{err_cross_global}
\end{figure}

The results provided in \cref{plots_global_post} and \cref{plots_global_post_std} assume no knowledge of bathymetries other than the PCGA estimation. In some practical applications, we may have some additional knowledge of the bathymetry. That is, after an original survey or estimation, we might gain new information about the bathymetries at a limited number of cross sections in the domain of interest, instead of no available cross section data (the case in \cref{plots_global_post} and \cref{plots_global_post_std}) or the complete data over the entire domain (the case in \cref{error_global} or \cref{plots_global_low} and \cref{plots_global_high}). The results shown here may be useful when due to time or computational resource constraints, we do not want to perform a new data assimilation or inversion using the new bathymetry information. We rely on our originally trained global solvers only. Here, we consider a number of cases where the bathymetry at 10, 25, 50, 100, 150, 200, 250, 300, 350, and 450 neighboring cross sections along the river are chosen. \Cref{loc_section} shows an example of such cases where we have the measured bathymetry of the first 150 cross sections (top left region) while for the rest of the river domain we only have the posterior mean estimate (and thus the discontinuity at the 150th cross section). 

Here, we perform an experiment, where, for any of our datapoints in the train, validation, or test datasets, we provide the measured bathymetries at any of these (10--450) cross sections along with the mean values (of the posterior) of bathymetries at other cross sections to the forward solver, and evaluate the performance of different solvers in predicting the flow velocities. In particular, if the first $S$ cross sections are measured, we calculate the RMSEs in the form of
\begin{equation}
\text{RMSE}= \sqrt{\frac{\sum_{j,k} \left( g_{j,k}(\hat{x}_{j})-y_{j,k} \right)^2}{N}}
\label{part_RMSE}
\end{equation}
where $g_{j,k}(\hat{x}_{j})$ is the $k$-th component of the $j$-th predicted flow velocity (the flow velocity prediction of the forward solver for the $j$-th member in the dataset), $y_{j,k}$ is the $k$-th component of the $j$-th (true) flow velocity for each data set member $j$ (see \cref{latent}), $N$ is the total number of points at which the error has been calculated (see \cref{latent}), and $\hat{x}_{j}$ is the bathymetry of the $j$-th dataset member input to the forward solvers (and $\hat{x}_{j,k}$ is its $k$-th component), defined as
\begin{eqnarray*}
\hat{x}_{j,k}&=& x_{j,k}, \mbox{ if } k \mbox{ is a point in the first } S \mbox{ cross sections, while } \\
\hat{x}_{j,k}&=& \bar{x}_{k} ,
\end{eqnarray*}
in which $x_{j,k}$ is  $k$-th component of the true bathymetry (of the $j$-th member in the dataset) and $\bar{x}_{k}$ is the  $k$-th component of the posterior mean value.

\begin{figure}[htbp]
\centering
\hspace{-1.2cm}
\begin{subfigure}{.325\textwidth}
\centering
\includegraphics[width=1.2\linewidth]{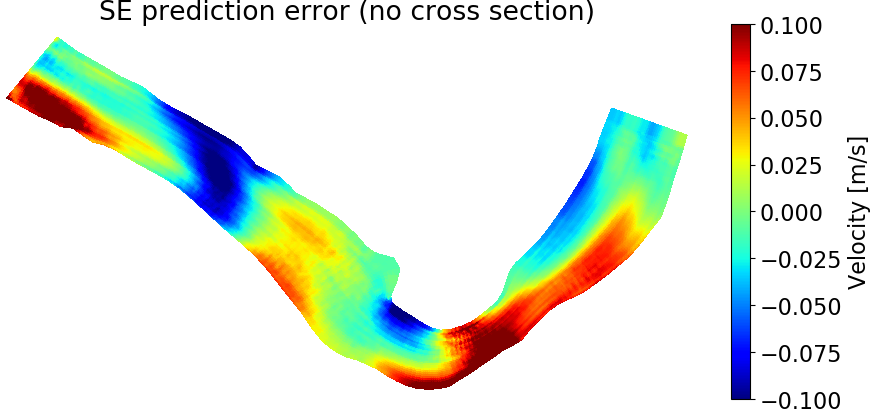}
\caption{No cross section}
\label{no_cross}
\end{subfigure}
\hspace{-0.57cm}
\begin{subfigure}{.325\textwidth}
\centering
\includegraphics[width=1.2\linewidth]{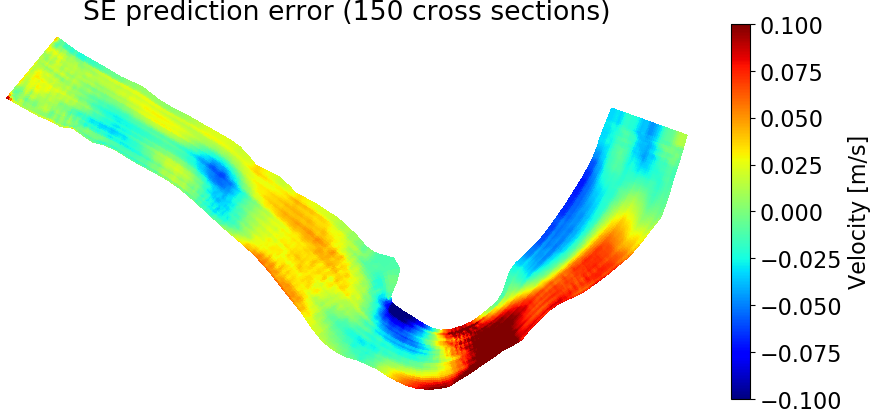}
\caption{150 cross sections}
\label{150_cross}
\end{subfigure}
\hspace{-0.57cm}
\begin{subfigure}{.325\textwidth}
\centering
\includegraphics[width=1.2\linewidth]{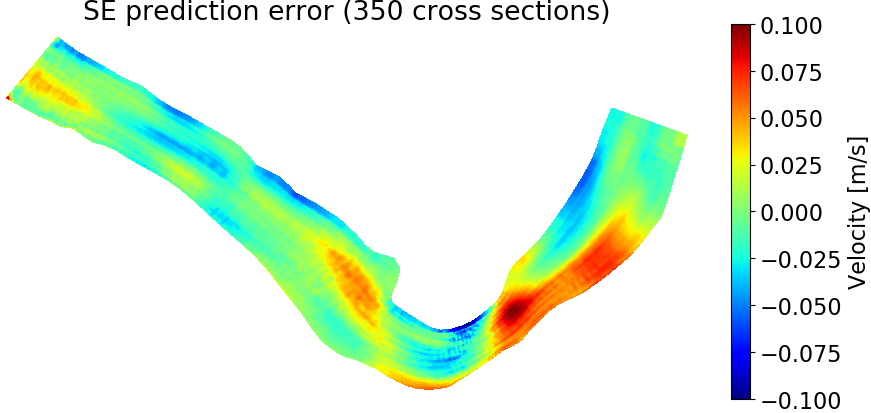}
\caption{350 cross sections}
\label{350_cross}
\end{subfigure}
\caption{Improvement in the flow velocity prediction error when 150 (middle) and 350 (right) cross section measurements are available compared to the case that no bathymetry measurement is available (left), for the SE approach. The additional input of the bathymetry measurements has improved the accuracy compared to no-bathymetry measurement case.}
\label{plots_global_pca_section}
\end{figure}

\Cref{err_cross} shows the errors in the prediction of the flow velocity in easting direction when measurements at different number of cross sections are provided as inputs to the DNNs. We observe that incorporating cross section bathymetry measurements can reduce the error significantly compared to the case where no knowledge of the bathymetry measurement is incorporated into the flow velocity prediction other than the mean of the PCGA estimation (leftmost points in the figure). We also observe, consistent with our previous results, that SVE and SE outperform PCA-DNN in all cases (all number of cross sections). \Cref{plots_global_pca_section} shows an example of the improvement in the flow velocity prediction for SE using measurements at 150 and 350 cross sections, for the same datapoint that the results in \cref{plots_global_low} (and the bathymetry in \cref{loc_section}) were based on. Note that an important feature of the solvers is that although our input to the solvers are not physically realistic, due to the jump at the locations where the segments with the bathymetry measurements meet the sections with no bathymetry measurements (see \cref{loc_section}), the solvers predict physically realistic flow velocities (\cref{150_cross} and \cref{350_cross}).

\section{Local solver}
\label{result_local}

\begin{figure}[htbp]
%\vspace{-1cm}
\centering
%\hspace{-1.2cm}
\begin{subfigure}{.49\textwidth}
%\vspace{-4cm}
\centering
\includegraphics[width=0.95\linewidth]{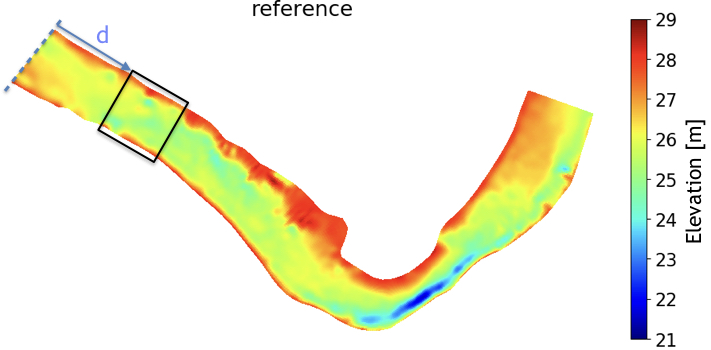}
\caption{}
\label{d_local}
\end{subfigure}
%\vspace{0.57cm}
\begin{subfigure}{.49\textwidth}
\vspace{0.72cm}
\centering
\includegraphics[width=1.05\linewidth]{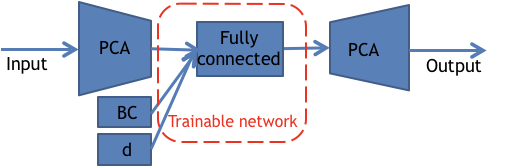}
\caption{}
\label{pca-dnn_sketch_local}
\end{subfigure}
\caption{(a) The distance $d$ used in the local solver. By providing this information to the network, we ensure that the network accounts for the location of any river segment provided to it as an input. (b) Schematic of the PCA-DNN as a local solver. The architecture is similar to the global solver case, except, the distance $d$ of any segment to the upstream is also provided as an additional input.}
\label{local_stuff}
\end{figure}

In this section, we discuss application of deep learning on flow velocity estimation when we input small sections of the river to the network and predict velocities only for that section. By dividing a river profile into smaller segments, the local solver has access to a significantly larger dataset size compared to the global solver, making it suitable for situations where due to scarcity of the data, applying a DNN to the high-dimensional input/output of the whole river leads to overfitting~\cite{PCA_DNN_Hojat}. Furthermore, in situations where we are interested in predicting flow velocity of a section of a river instead of the whole river, having a local solver is useful. The network structures in the local case are similar to the global case, however, we also provide the location of the cross section as another input to the network, that is, the distance of the section to the inlet of the simulation domain (top left edge of the reach). This distance is shown in \cref{d_local}. 

\Cref{pca-dnn_sketch_local} shows the PCA-DNN architecture as the local solver where the distance $d$ is added as an additional input; for SE and SVE also the process is similar. The input in this figure is the river profile for a section of the river, here of size $41\times 16$ (about 0.032 of the riverine domain), and the output is the flow velocity for the same section with the same size. We have also assumed the same latent space dimension of 50 as the global solver. Therefore, the input of DNN part of the architecture has 53 components (50 for the low-dimensional bathymetry representation, 2 for the two BCs, and one for the parameter $d$). For cases that the size of the desired section is larger than the window size of the input/output, we use the local forward solver for smaller subsets of that section with the size equal to the input and output of the network and then use the average of predictions by different sections for the estimated flow velocities. For instance, if the size of our section is $41\times 18$, located at $d$, we can use the local solver either (a) three times on different $41\times 16$ segments (size of the input/output of the network); located at distances of $d$, $d+1$, and $d+2$ and use the mean of predictions for grid points which are common in 2 or 3 of the small segments; or (b) twice on segments that do not have any overlap, such as the one located at $d$ and the one located at $d+16$.

\Cref{error_local} compares the errors of different local solvers when they are used for prediction of the flow velocities in easting direction of the Savannah river. The train, validation, and test dataset used here are the same as the ones used in the global solver case; this will lead to a dataset size of $4500 \times 486 =$ 2,187,000 for train/validation, where 4,500 is the number of riverine domains (as mentioned before while discussing the global solvers) and $486$ is the number of small segments available in each riverine profile. An important observation in the results presented in \cref{error_local} is the significantly larger error of the linear model, in contrast to what was observed with the global solver in \cref{perform_glob} (see also \href{run:./SI.pdf}{Supplementary Information} file). The errors in PCA-DNN, SE, and SVE are larger than the global solver cases (see \href{run:./SI.pdf}{Supplementary Information} file). This is likely because the local solver needs to learn a more complicated function compared to the global solver, since the input-to-output map in the case of local solver should capture the dynamics of all windows (of size $41\times 16$) located at different distances to the upstream, separately. This map can be very different, for instance, in the inlet/outlet region compared to the bend of the river. The large errors can also be due to the averaging performed when there are cross sections with multiple windows. 

\begin{table}[htbp]
    \centering
    \begin{tabular}{lllll}
        \toprule
        \multirow{2}{*}{Error} & \multicolumn{4}{c}{Fast forward solver}\\
        \cmidrule{2-4} \cmidrule{5-5}
        {} & PCA-DNN & PCA with linear map & SE & SVE \\
        \midrule
        Train set RMSE [m/s]   & 0.0468 & 0.1568 & {\bf 0.0396} & 0.0430\\
        Validation set RMSE [m/s] & 0.0503 & 0.1579 & 0.0466 & {\bf 0.0460}\\
        Test set RMSE [m/s] & 0.0517 & 0.1686 &  {\bf 0.0457} & 0.0459\\
        \bottomrule
    \end{tabular}
    \caption{Comparison between the error of different local solvers in the prediction of the flow velocities.}
    \label{error_local}
\end{table}

\begin{figure}[htbp]
\centering
\begin{subfigure}{.49\textwidth}
\centering
\includegraphics[width=1.0\linewidth]{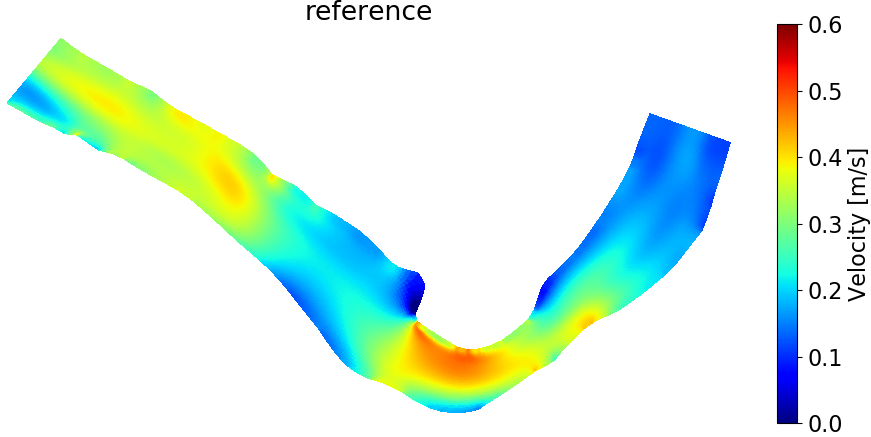}
%\caption{}
\label{ab_initiomodel}
\end{subfigure}
\begin{subfigure}{.49\textwidth}
\centering
\includegraphics[width=1.0\linewidth]{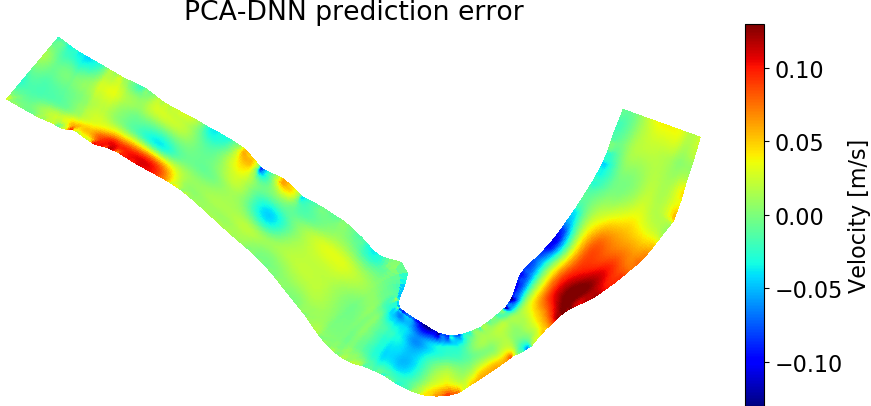}
%\caption{}
\label{Hollandmodel}
\end{subfigure}
\begin{subfigure}{.49\textwidth}
\centering
\includegraphics[width=1.0\linewidth]{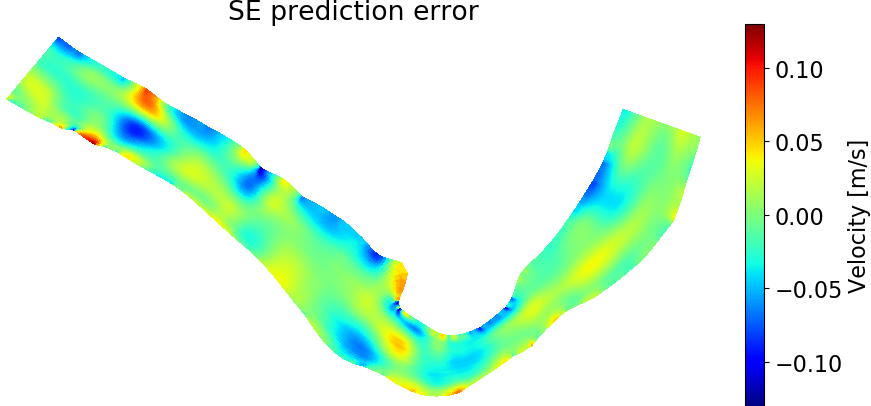}
%\caption{}
\label{ab_initiomodel}
\end{subfigure}
\begin{subfigure}{.49\textwidth}
\centering
\includegraphics[width=1.0\linewidth]{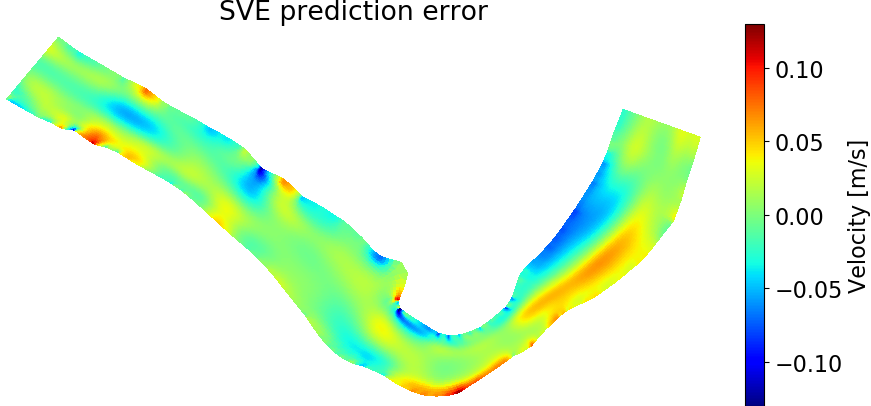}
%\caption{}
\label{Hollandmodel}
\end{subfigure}
\caption{Examples of error in the predicted flow velocities of different local solvers for easting direction. The BCs are the same as the ones used in \cref{plots_global_low}.}
\label{plots_local}
\end{figure}

\begin{figure}[htbp]
\centering
\begin{subfigure}{.49\textwidth}
\centering
\includegraphics[width=1.0\linewidth]{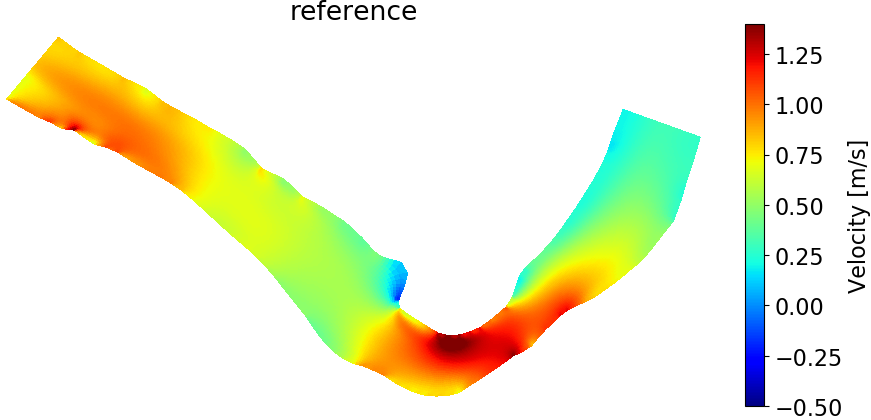}
%\caption{}
\label{ab_initiomodel}
\end{subfigure}
\begin{subfigure}{.49\textwidth}
\centering
\includegraphics[width=1.0\linewidth]{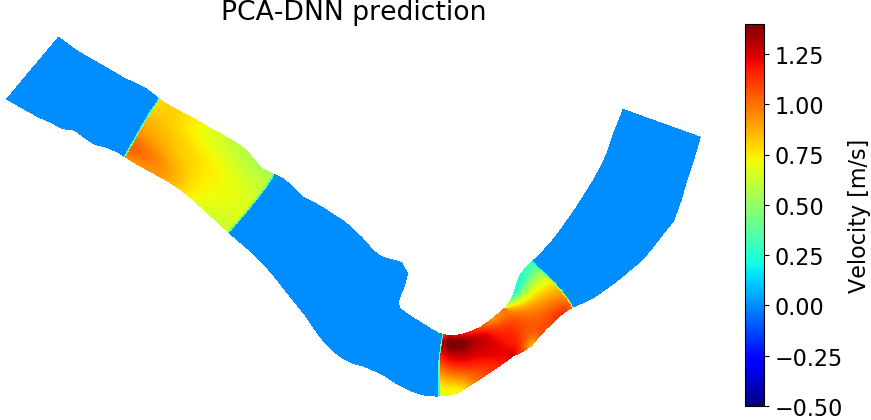}
%\caption{}
\label{Hollandmodel}
\end{subfigure}
\begin{subfigure}{.49\textwidth}
\centering
\includegraphics[width=1.0\linewidth]{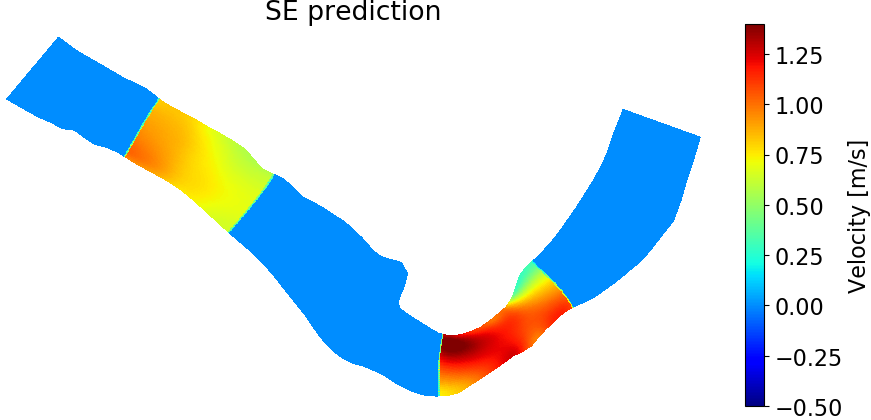}
%\caption{}
\label{ab_initiomodel}
\end{subfigure}
\begin{subfigure}{.49\textwidth}
\centering
\includegraphics[width=1.0\linewidth]{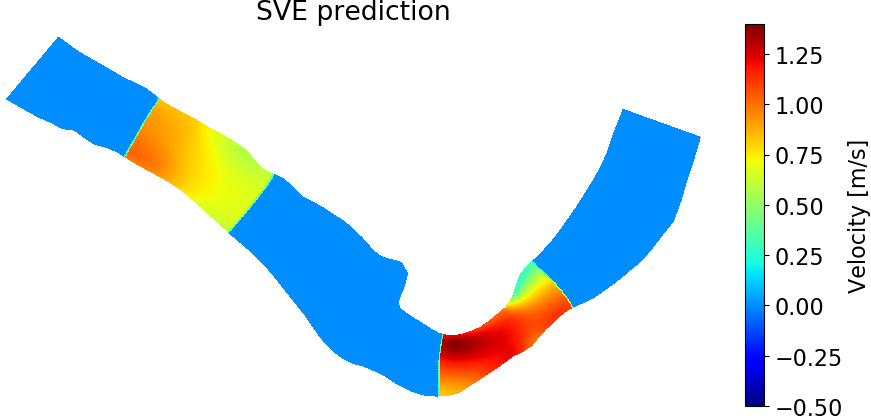}
%\caption{}
\label{Hollandmodel}
\end{subfigure}
\caption{Examples of flow velocities prediction of different local solvers for easting direction. The BCs are the same as the ones used in \cref{plots_global_high}.}
\label{plots_local_seg}
\end{figure}

For the results in this section, we used a fully connected DNN with 6 hidden layers in the PCA-DNN and an incremental-PCA algorithm~\cite{IPCA} to perform the PCA, due to the large size of the dataset. The SE and SVE also consist of 5 hidden fully connected layers in either encoder or decoder. For all three methods we have tried other network architectures as well (1--7 hidden layers). We have also used batch normalization for all networks, in contrast to the global solver case which showed better performance without the batch normalization. Note that here since the dimension of input/output is smaller than the global case, we have used fully connected network for SE/SVE instead of convolutional layers. \Cref{plots_local} shows the errors in prediction of the flow velocity in easting direction for the whole riverine domain using different local solvers for the same BCs that the result of \cref{plots_global_low} was based on. \Cref{plots_local_seg} shows the prediction of flow velocity in easting direction for two different riverine segments of size 80 (80 cross sections) and 120 for the same BCs that the result of \cref{plots_global_high} was based on.

\section{Conclusion}
\label{conclusion}

In this work, we have presented a framework for fast prediction of the riverine flow velocities with variable boundary conditions and bathymetries. The training of all the presented methods can be performed on common personal computers without access to GPU and high-performance computing resources. More importantly, once the networks are trained, the predictions can be done in a few seconds, making online flow velocity estimations possible. Our results show the computational efficiency about three orders of magnitude faster than common SWE solvers (such as AdH)---for instance, a single prediction of the flow velocity for given bathymetries and BCs takes about 15--20 min using AdH and only one second using our approaches on the 48 core workstation used in this paper (see \cref{perform_glob}).

The combination of the PCGA and our fast solvers provides a valuable tool that can be used even when the riverine bathymetry profiles are not a-priori available. That is, we do not need to measure riverbed profiles when training the network and designing the fast, reduced-order solver (offline stage). More importantly, even for the online prediction stage, we can predict distribution of flow velocities from the posterior distribution of the PCGA, without access to updated bathymetry observations. Furthermore, any additional dense/sparse measurement of bathymetries can be directly incorporated into the flow velocity prediction, anytime that they become available, without requiring any additional training.

While all of the presented global solvers are capable of providing reasonable prediction of the flow velocities, the better performance of SE and SVE methods implies that there are non-linear features present in the data that linear models or partially-linear models such as PCA-DNN may not be able to capture accurately within the available computational limitations. Good performance of all solvers on the test set is essential for generalization of the forward solvers. Note that neither the bathymetries or the BCs in the test set were present in the train/validation set during the training processes. In general, global solvers outperform their local solver equivalents. However, the local solvers allow prediction of flow velocities for segments of arbitrary lengths along the river which can be useful in cases where we are interested in solutions only in that segment. Furthermore, in cases where we do not have sufficient data to train a network for the entire river, due for instance to high computational cost of generating data for long rivers, the segmentation process used in local solvers help to overcome overfitting by reducing dimension of data and increasing the size of the dataset. 

In \cref{error_dist} we observe that the errors are almost uniformly distributed among most of the BCs, leading to smaller relative errors for large BCs, i.e., large discharges. Thus, the proposed solvers will be beneficial for high-flow situations. We can control the distribution of errors by changing the loss function from RMSEs, for example, to relative RMSE, or RMSEs with higher weight for large BCs to ensure that the prediction accuracy is higher when the risk of water overflow is high. Similarly, we can increase the weight of the datapoints with small BCs to reduce their errors (\cref{error_dist}). The more complex dynamics of these datapoints (\cref{worst_small}) can also imply the need for a latent space with larger dimension, and consequently a larger training set (to avoid overfitting).

The bathymetries provided to our DNNs were obtained by augmented posterior distribution of the PCGA (the Gaussian kernel and the scaling factor). In future, we will study incorporation of more complex and general distributions of bathymetries into our training sets, thus generalizing the forward solver to a larger class of bathymetries. Another assumption in this work is that the geometry of the reach was fixed. Obvious extensions include updating the methodology to allow for significant changes in the lateral geometry (due for example to bank overflow) and extending the approach to allow application to multiple classes of river.

Finally, for the cases with the bathymetry measurement at a limited number of cross sections, we have studied the cases where the measurement at first $S$ cross sections is available. However, depending on the uncertainty observed in the flow velocity predictions, one should be able to find specific cross sections that are more influential in reducing the error in predictions in order to guide collection of new bathymetry data. For instance, in Figure \ref{plots_global_post_std}, we find regions of high uncertainty near the entrance of the river. This would fit naturally into design of experiments based approaches~\cite{DoE} and is an area future interest.

\section{Acknowledgements}

The PCGA codes can be found in the \href{https://github.com/jonghyunharrylee/pyPCGA}{pyPCGA} github repository, and the DNN codes can be found in the \href{https://github.com/moji1369/DNN-based-fast-solver-of-SWEs}{DNN-SWEs} github repository. This research was supported by the U.S. Department of Energy, Office of Advanced Scientific Computing Research under the Collaboratory on Mathematics and Physics-Informed Learning Machines for Multiscale and Multiphysics Problems (PhILMs) project, PhILMS grant DE-SC0019453. Jonghyun Lee was supported by Hawai'i Experimental Program to Stimulate Competitive Research (EPSCoR) provided by the National Science Foundation Research Infrastructure Improvement (RII) Track-1: 'Ike Wai: Securing Hawai'i's Water Future Award OIA \#1557349. This work was also supported by an appointment to the Faculty and Postdoctoral Fellow Research Participation Program at the U.S.\ Engineer Research and Development Center, Coastal and Hydraulics Laboratory administered by the Oak Ridge Institute for Science and Education through an inter-agency agreement between the U.S. Department of Energy and ERDC. The Chief of Engineers has granted permission for the publication of this work.
\printbibliography

\end{document}